\documentclass[journal]{IEEEtran}
\usepackage{graphicx}
\usepackage{subfigure}
\usepackage{amsmath,amsfonts,amssymb}
\usepackage{algorithm}
\usepackage{algorithmic}
\usepackage{booktabs}
\usepackage{cite}
\usepackage{color}

\begin{document}

\title{3D UAV Trajectory Planning for IoT Data Collection via Matrix-Based Evolutionary Computation}

\author{Pei-Fa Sun, Yujae Song,~\IEEEmembership{Member,~IEEE,} Kang-Yu Gao, Yu-Kai Wang, Changjun Zhou,~\IEEEmembership{Member,~IEEE,} Sang-Woon Jeon,~\IEEEmembership{Senior Member,~IEEE,} and~Jun~Zhang,~\IEEEmembership{Fellow,~IEEE}
\thanks{P.-F. Sun, K.-Y. Gao, and Y.-K. Wang are with the Department of Electrical and Electronic Engineering, Hanyang University, ERICA, Ansan, South Korea (e-mail: \{{pfsun, gaokangyu, ykwang}\}@hanyang.ac.kr).}
\thanks{Y. Song is with the Department of Robotics Engineering, Yeungnam University, Gyeongsan, South Korea (e-mail: yjsong@yu.ac.kr)}
\thanks{Changjun Zhou is with the School of Computer Science and Technology, Zhejiang Normal University, Jinhua, China (e-mail: zhouchangjun@zjnu.edu.cn).}
\thanks{S.-W. Jeon is with the School of Computer Science and Technology, Zhejiang Normal University, Jinhua, China and the Department of Electrical and Electronic Engineering, Hanyang University, ERICA, Ansan, South Korea (e-mail: sangwoonjeon@hanyang.ac.kr).}
\thanks{J. Zhang is with the Hanyang University, ERICA, Ansan, South Korea, with the Nankai University, Tianjin, China, and also with the Victoria University, Melbourne, Australia (e-mail: junzhang@ieee.org).}
\thanks{\emph{Corresponding authors: Sang-Woon Jeon and Jun Zhang.}}}
% \markboth{Journal of \LaTeX\ Class Files,~Vol.~14, No.~8, August~2015}%
% {Shell \MakeLowercase{\textit{et al.}}: Bare Demo of IEEEtran.cls for IEEE Journals}

\maketitle

\begin{abstract}
UAVs are increasingly becoming vital tools in various wireless communication applications including internet of things (IoT) and sensor networks, thanks to their rapid and agile non-terrestrial mobility. Despite recent research, planning three-dimensional (3D) UAV trajectories over a continuous temporal-spatial domain remains challenging due to the need to solve computationally intensive optimization problems. In this paper, we study UAV-assisted IoT data collection aimed at minimizing total energy consumption while accounting for the UAV's physical capabilities, the heterogeneous data demands of IoT nodes, and 3D terrain. We propose a matrix-based differential evolution with constraint handling (MDE-CH), a computation-efficient evolutionary algorithm designed to address non-convex constrained optimization problems with several different types of constraints. Numerical evaluations demonstrate that the proposed MDE-CH algorithm provides a continuous 3D temporal--spatial UAV trajectory capable of efficiently minimizing energy consumption under various practical constraints and outperforms the conventional fly--hover--fly model for both two-dimensional (2D) and 3D trajectory planning.
\end{abstract}


\IEEEpeerreviewmaketitle

\section{Introduction}
% [UAV general descirption]
% \IEEEPARstart{U}{nmanned} aerial vehicle (UAV) has played a more and more important role due to its dynamics and swift mobility characteristic, utilizing in many fields, such as search and rescue \cite{Wu2023rescue}, wildfire detection \cite{Bushnaq2021wildfire}, and especially wireless communication \cite{zeng2019accessing, ullah2020cognition, geraci2022will}.
% Especially in wireless communication areas, UAVs can effectively build the line-of-sight (LoS) channel with ground nodes (GNs) to ensure high-quality of service (QoS) communication. So far, there have been many UAV-aided applications related to wireless communication, e.g., data collection from the Internet of Things (IoTs) \cite{hu2020aoi} \cite{wang2021trajectory}, mobile edge computing and secure communications \cite{zhou2018computation,zhou2019secure,lu2021resource,zhou2023priority}.
% Thus, studying UAV-enabled wireless communications is vital to give powerful underlying support to these applications.

\IEEEPARstart{I}{n} recent years, unmanned aerial vehicles (UAVs) have played a vital role in wireless communication networks, extending network coverage and enhancing service quality due to their ability to operate in three-dimensional (3D) space \cite{zeng2019accessing, ullah2020cognition, geraci2022will}. The integration of UAVs into wireless networks provides unparalleled mobility and real-time data collection capabilities, proving to be invaluable in various scenarios, including disaster response, wildlife conservation, infrastructure inspection, and jamming \cite{Yujae2020, Wu2023rescue, Bushnaq2021wildfire, Haroon2023}.

% [Advantage of UAV utilization in sensor networks]
As one of the representative applications utilizing UAVs in wireless networks, UAV-assisted sensor data collection has garnered substantial attention among researchers. In traditional sensor network scenarios, where sensors transmit their sensing data to control centers or cluster heads, communication performance deteriorates as the geographical distance between two nodes increases, primarily due to losses incurred in electromagnetic wave propagation. To address this challenge, the deployment of UAVs for data acquisition from distributed sensors has emerged as a promising solution. The key issue in the UAV-assisted sensor data gathering scenario, wherein UAVs play a role in collecting data from sensors distributed in specific geographical areas, is to design efficient trajectories for UAVs.

% [Existing work on trajectory optimiztion of UAV ]
Over the past few years, many studies on the UAV trajectory optimization have been conducted under various UAV-enabled wireless communication network scenarios~\cite{guo2019uav,hu2020aoi,Minho2020, na2020uav, samir2019uav, li20213d, wang2021trajectory,zhu2021uav,li2022dynamic, Kang2024}. 
The study in \cite{guo2019uav} conducted joint optimization of task offloading decision-making, bit allocation during transmission, and UAV trajectory in the UAV-enhanced edge network, considering the limited battery capacity of internet of things (IoT) mobile devices and UAV energy budget. 
In \cite{hu2020aoi}, a dynamic programming-based algorithm minimized the average age of information of data collected from all sensors by optimizing both the time required for energy harvesting and the UAV trajectory in the UAV-assisted wireless powered IoT system. 
In \cite{Minho2020}, a set of optimal curvature-constrained closed trajectories for multiple UAVs was established. 
The research in \cite{na2020uav} focused on a UAV-supported clustered non-orthogonal multiple access system, investigating a joint algorithm that optimizes both the UAV trajectory and subslot duration allocation. %The goal was to maximize the uplink average achievable sum rate of IoT terminals while ensuring that the uplink achievable sum rate and the UAV mobility constraints are met.
The research in \cite{samir2019uav} addressed how many time-constrained IoT devices a UAV can accommodate when integrated into existing wireless networks by jointly optimizing UAV trajectory and radio resource allocation for multiple access techniques, aiming to ensure each IoT device meets its individual target data upload deadline.
The authors in \cite{li20213d} presented a practical 3D trajectory optimization algorithm, for which a new energy consumption model for electric quad-rotor UAVs was developed, considering the current and voltage of four battery-powered brushless motors.
In \cite{wang2021trajectory}, a deep reinforcement learning (DRL) approach was used for UAV trajectory design within a real 3D urban environment, optimizing trajectory to minimize data collection time without sacrificing practical throughput and flight movement constraints.
Similar to the approach in \cite{wang2021trajectory}, the authors in \cite{zhu2021uav} employed a DRL algorithm, focusing on jointly selecting cluster heads from clusters and determining the visiting order of the UAV to the selected cluster heads within large-scale wireless sensor networks.
Furthermore, in \cite{li2022dynamic}, a two-level DRL framework was utilized to address diverse practical scenarios in UAV trajectory optimization for data collection in wireless sensor networks, comprising a first-level deep neural network (DNN) modeling the environment and a second-level deep Q-learning network for online trajectory planning based on environmental features from the DNN.
In \cite{Kang2024}, energy-efficient data aggregation and collection for multi-UAV-enabled IoT networks was considered by integrating both multi-hop and UAV routing.

In the above literature survey on UAV trajectory design, it becomes evident that UAV energy consumption emerges as a critical performance metric for optimizing UAV trajectories.
Consequently, researchers have made additional efforts to precisely model the energy consumption of various UAV types.
The authors in \cite{zeng2019energy} introduced a generalized propulsion energy consumption model (PECM), which is a mathematical expression that establishes the relationship between UAV power and the varying UAV speed over time.
However, PECM presented in \cite{zeng2019energy} solely considered the horizontal speed, neglecting vertical speed, thereby limiting its applicability to two-dimensional (2D) environments.
In \cite{mei20223d}, the authors considered both the vertical speed and position of the UAV, extending the previously 2D PECM to a 3D framework.
In \cite{ding20203d}, the study also aimed to expand the existing PECM from 2D to 3D, considering not only vertical speed but also enhancing the practicality of the PECM for quad-rotor UAVs.
The authors in \cite{dai2022energy} presented a 3D PECM in which acceleration plays a crucial role in determining UAV power. 
Furthermore, the work in \cite{dai2023wind} considered the real effects of wind, providing a solid foundation for practical applications.
Table~\ref{tab:comparison_work} summarizes key features of the existing works \cite{zeng2019energy, mei20223d, ding20203d, dai2022energy, dai2023wind} using various PECMs for UAV trajectory planning.
\begin{table*}[t!]
\caption{Summary of the key features of existing works using various PECMs for UAV trajectory planning.}\label{tab:comparison_work}
\centering
\setlength{\tabcolsep}{1.5mm}{
\begin{tabular}{lllllll}
\toprule
Article  &Dim&Trajectory Rep. & Energy Model & 3D Terrain & Objective &Optimization Method  \\ \midrule
\cite{zeng2019energy}&2D  &Discrete Point &2D PECM&  No &Min. Energy Consumption& SCA and Convex Optimization\\
\cite{mei20223d} &3D & Discrete Point & 3D PECM & No & Min. Energy Consumption & Deep Reinforcement Learning\\
\cite{ding20203d}&3D&Discrete Point&3D PECM (Quad-rotor) & No &Max. Fair Throughput& Deep Reinforcement Learning\\
%\textcolor{red}{\cite{li20213d}}&2/3D&Discrete Point&Electrical Quad-rotor Model&No&Min. Energy/Time&Sequential Quadratic Programming etc.\\
\cite{dai2022energy}&2D&Discrete Point& 3D PECM (with Acceleration) & No & Max. Energy Efficiency &AO and Dinkelbach's Method with SCA\\
\cite{dai2023wind}&3D &Discrete Point&3D PECM (with Wind)& Yes &Max. Minimum Data Rate& SCA and Dinkelbach’s Method\\
\bottomrule
\end{tabular}}
\end{table*}
In Table~\ref{tab:comparison_work}, `Trajectory Rep.', `SCA', `AO' stand for trajectory representation, successive convex approximation, and the alternative optimization, respectively. Table~\ref{tab:comparison_work} reveals two notable limitations in the current body of research. 
Firstly, the existing studies commonly focus on linking discrete points when designing UAV trajectories. This necessitates a greater number of points for smoother UAV trajectories, which can significantly increase computational demands and the complexity of optimization tasks.
Secondly, the existing literature often neglects the impact of 3D terrain, a critical factor for UAV operational safety in real-world applications. 
Accounting for terrain elevations converts an optimization problem into a non-convex one, requiring sophisticated reformulations of conventional optimization approaches.

To address these challenges, this paper presents the following key contributions:
\begin{itemize}
    \item We have formulated an optimization problem to plan a continuous UAV trajectory aimed at minimizing the total energy consumption necessary for mission completion (e.g., data gathering from distributed sensors) while taking into account various constraints such as the UAV's physical capabilities, diverse communication requirements for each sensor, and the need for avoiding collisions with 3D terrain.
    \item We adopt B\'{e}zier curves to efficiently generate continuous UAV trajectories that can significantly reduce the number of decision variables in the optimization problem, facilitating low-complexity real-time solutions.
    \item To solve the formulated optimization problem, we propose a matrix-based differential evolution with constraint handling (MDE-CH), which is a computation-efficient evolutionary algorithm designed to address non-convex optimization problems that deal with various types of constraints. 
    \item Through comprehensive numerical simulations, we demonstrate the performances of the proposed MDE-CH algorithm across various network scenarios, including 3D terrain presence and variations in communication throughput requirements. Our simulations validate the effectiveness of the MDE-CH algorithm, producing UAV trajectories that exhibit consistent smoothness, thereby confirming their operational viability and adaptability for real-world implementation.
\end{itemize}

The rest of the paper is organized as follows. Section \ref{sec:Problem_formulation} provides a problem formulation including the system model and the UAV energy consumption and channel model for data transmission and reception. In Section \ref{sec:Optimization_for_Continuous_UAV_Trajectory}, UAV trajectory optimization over a continuous temporal--spatial domain is described. In Section \ref{sec:MDE}, we propose an MDE-CH algorithm to construct 3D UAV trajectory.
Performance evaluation is conducted in Section \ref{sec:simulation} and concluding remarks are presented in Section \ref{sec:conclusion}.

\section{Problem Formulation}\label{sec:Problem_formulation}
In this section, we first introduce the system model including the energy consumption model for UAV traveling and channel model for data gathering. Then we present the UAV trajectory optimization problem considered in this paper.

\subsection{System Model}\label{sec:system_model}
\begin{figure}[t!]
    \centering
    \includegraphics[width=1\linewidth]{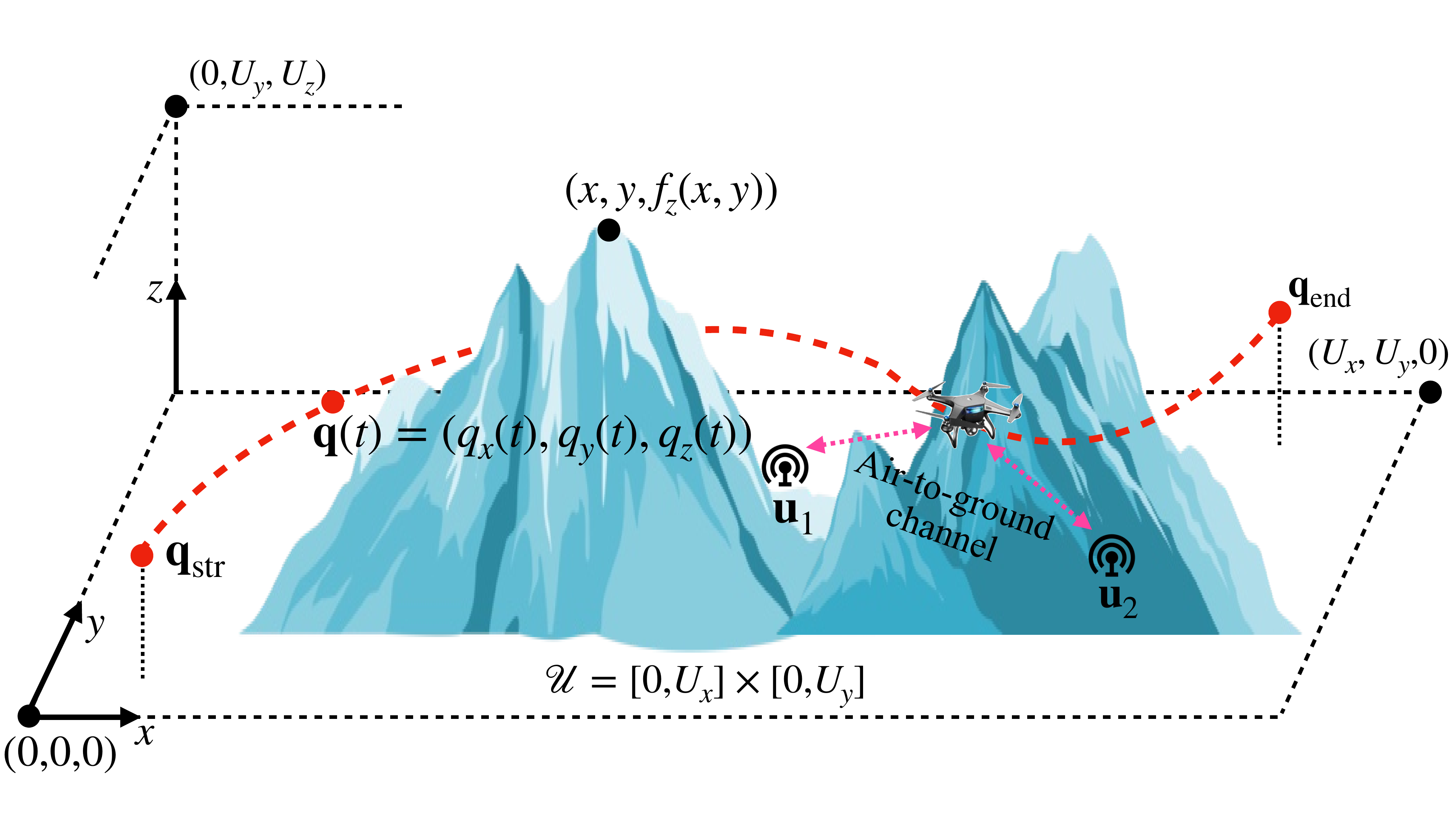}
    \caption{3D UAV trajectory planning for IoT data collection.}
    \label{fig:system_model}
\end{figure}

We consider a UAV-aided IoT data collection depicted in Fig. \ref{fig:system_model} in which a single UAV is employed to communicate with $K$ ground nodes (GNs). 
Let $\mathcal{U}=[0, U_{x}]\times [0, U_{y}]$ be the 2D network area. This paper considers a 3D space that incorporates arbitrary terrains and physical structures. For such purpose, denote a mapping function $f_z: \mathcal{U}\rightarrow \mathbb{R}$ such that $f_z(x, y)$ returns the ground altitude at the 2D position $(x, y)\in \mathcal{U}$ including artificial structures such as buildings.
Let $[u_{k,x},u_{k,y}]\in \mathcal{U}$ be the 2D position of GN $k$. Then the corresponding 3D position of GN $k$ is given by  
\begin{align} \label{eq:u_k}
\mathbf{u}_k=\left[u_{k,x},u_{k,y}, f_z(u_{k,x}, u_{k,y})\right].
\end{align}
That is, the position of GN $k$ along the z-axis given by $u_{k,z}=f_z(u_{k,x}, u_{k,y})$.
Denote $\mathbf{q}_{\operatorname{str}}=[q_{\operatorname{str},x},q_{\operatorname{str},y},q_{\operatorname{str},z}]$ and $\mathbf{q}_{\operatorname{end}}=[q_{\operatorname{end},x},q_{\operatorname{end},y},q_{\operatorname{end},z}]$ as the UAV's start position and end position, respectively. To be valid $[q_{\operatorname{str},x},q_{\operatorname{str},y}]\in\mathcal{U}$ and $q_{\operatorname{str},z}\geq f_z(q_{\operatorname{str},x},q_{\operatorname{str},y})$ should be satisfied. Similarly,
$[q_{\operatorname{end},x},q_{\operatorname{end},y}]\in\mathcal{U}$ and $q_{\operatorname{end},z}\geq f_z(q_{\operatorname{end},x},q_{\operatorname{end},y})$ should be satisfied.

The aim of the UAV is to collect data from each of $K$ GNs, initially starting from $\mathbf{q}_{\operatorname{str}}$ and returning to $\mathbf{q}_{\operatorname{end}}$ after data gathering.
Let 
\begin{align}
\mathbf{q}(t)=[q_x(t),q_y(t),q_z(t)]
\end{align}
be the 3D position of the UAV at time $t\in[0,T]$, where $T$ denotes the task completion time which is bounded by the maximum allowable time for operating the UAV denoted by $T_{\operatorname{max}}$, such that ${T}  \le  {T_{\max }}$.
Then by sweeping $\mathbf{q}(t)$ from $t=0$ to $t=T$, the temporal--spatial trajectory of the UAV can be specified by $\{\mathbf{q}(t)\}_{t\in[0,T]}$. To be valid $\{\mathbf{q}(t)\}_{t\in[0,T]}$, the following set of conditions should be satisfied:
\begin{align}
\begin{cases}
    \mathbf{q}(0)=\mathbf{q}_{\operatorname{str}},\\
\mathbf{q}(T)=\mathbf{q}_{\operatorname{end}},\\
[q_x(t),q_y(t)]\in \mathcal{U} {~~}\forall t\in(0,T),\\
{q_z(t)}\in[f_z(q_x(t),q_y(t))+\Delta_z, U_z] {~~}\forall t\in(0,T),
\end{cases}
\end{align}
where $\Delta_z>0$ represents the minimum distance that should be maintained from the ground altitude, and $U_z$ represents the maximum allowable altitude for the UAV.

In this paper, we consider two different types of UAV energy consumption. The first is the UAV traveling energy, which is determined by the 3D trajectory of the UAV $\{\mathbf{q}(t)\}_{t\in[0,T]}$. The second is the communication task energy, which is consumed for the transmission and reception of sensing data at each GN and the UAV. In this case, not only $\{\mathbf{q}(t)\}_{t\in[0,T]}$ but also other factors such as the required amount of data to be transmitted at each GN will affect the total communication energy. 
For the following two subsections, we formally state these two energy consumption models. 

\subsection{Energy Consumption Model for UAV Traveling}
%According to \textcolor{blue}{[XX]}, the total energy consumption of UAV consists of flying energy consumption $E_{\operatorname{fly}}$ and communication energy consumption $E_{\operatorname{com}}$. 
Before describing the UAV traveling energy model adopted in this paper, we first define the velocity, speed, and acceleration of the UAV, which affect the UAV traveling energy. 
%Let ${T}$ be the total time required for the UAV to complete its mission (e.g., gathering sensing data from multiple GNs), which will be considered as one of the optimization variables in this work. 
%Let $T_{\operatorname{max}}$ be the maximum allowable time for operating the UAV, such that $0 < {T} < {T_{\max }}$. %Then, we denote $u = \frac{t}{T}$ as time in the form of unit.  
%For any $u_\kappa < u_{\kappa+1}$ and $(u_{\kappa+1} - u_\kappa) \rightarrow 0$, we have
%\begin{equation}
%    \Delta u = u_{\kappa+1} - u_{\kappa} = \frac{u_{\kappa+1} - u_{\kappa}}{T} = \frac{\Delta t}{T}
%\end{equation}
% Define the time allocation as $\Delta t(u)$ in any very small segment $u + \Delta u$ ($\Delta u \rightarrow 0$) at the trajectory. For any $u_i, u_j \in (0,1)$, we have $\Delta t(u_i) = \Delta t(u_j)$. Thus, we use $\Delta t$ to denote $\Delta t(u)$.
% $\Delta t$ and $T$ need to follow 
% \begin{equation}\label{eq:time_T}
%     T = \int_{0}^{1} {\Delta t du},
% \end{equation}
% meaning we can calculate $\Delta t$ if $\Delta u$ is determined, namely $\Delta t = T \Delta u$.
The velocity of the UAV in meters per second (m/s) at time $t$ in the trajectory is given by
\begin{align}\label{eq:v_deri}
    \mathbf{v}(t) &= {\left[v_{x}(t), v_{y}(t), v_{z}(t)\right]}\nonumber\\
    & = \lim_{\Delta_t\to 0}\frac{\mathbf{q}(t+\Delta_t)-\mathbf{q}(t)}{ \Delta_t}.
\end{align}
Then, its speed is a scalar quantity of the velocity that is represented as  
\begin{equation}\label{eq:s_deri} 
    v(t) = \|\mathbf{v}(t)\|,
\end{equation}
where $\|\cdot\|$ is the norm of a vector.
Using \eqref{eq:v_deri}, we can express the acceleration $\mathbf{a}(t)$ in meters per second squared (m/{s$^2$}) as 
\begin{align}\label{eq:a}
    \mathbf{a}(t)  &= {\left[a_{x}(t), a_{y}(t), a_{z}(t)\right]}\nonumber\\
    &= \lim_{\Delta_t\to 0}\frac{\mathbf{v}(t+\Delta_t) - \mathbf{v}(t)}{\Delta_t}.
\end{align}
%We denote $\mathbf{q}(t) = \mathbf{B}\left(\frac{t}{T}\right)$ as the trajectory changing with time $t$ in the form of unity. 
%Fig.~\ref{fig:B_a_v} illustrates the relationship among location, velocity, and acceleration in the UAV trajectory. 
%The trajectory coordinate, velocity, and acceleration along the three dimensions ($x, y, z$) changing with time $t$ are given as
%\begin{align}
%    \begin{cases}
%        \mathbf{q}(t) = {\left[q_{x}(t), q_{y}(t), q_{z}(t)\right]},\\
%        \mathbf{v}(t) = {\left[v_{x}(t), v_{y}(t), v_{z}(t)\right]},\\
%        \mathbf{a}(t) = {\left[a_{x}(t), a_{y}(t), a_{z}(t)\right]}.
%    \end{cases}
%\end{align}
%\begin{figure}
%    \centering
%    \includegraphics[width=1\linewidth]{fig3/q_a_v.pdf}
%    \caption{The diagram of the location, velocity, and acceleration in the trajectory.}
%    \label{fig:B_a_v}
%\end{figure}

With \eqref{eq:v_deri} to \eqref{eq:a}, we consider the UAV traveling energy consumption model introduced in \cite{mei20223d}, which is an extended model from the 2D UAV velocity--power model \cite{zeng2019energy} to the 3D model.
In \cite{mei20223d}, the UAV traveling power consumption in watts ($W$) at time $t$  is calculated based on $\mathbf{v}(t)$ by 
\begin{multline}\label{energy_GPECM}
{P_{{\rm{fly}}}}\left(\mathbf{v}(t)\right) = \frac{1}{2}d_0\rho s A v_{xy}^3(t) + P_0\left(1+\frac{3 v_{xy}^2(t)}{U^2_{\operatorname{tip}}}\right)  \\
+P_1{\left(\sqrt{1+\frac{v_{xy}^4(t)}{4 v_0^4}} - \frac{v_{xy}^2(t)}{2 v_0^2}\right)}^{\frac{1}{2}}
+ P_2 \left|v_z(t)\right|,
\end{multline}
where $P_0$ and $P_1$ are the constant blade profile power and the induced power in hovering status, respectively; $P_2$ is the constant descending/ascending power; $U_{\operatorname{tip}}$ is the tip speed of the rotor blade; $v_0$ is the mean rotor-induced speed in hovering status; $s$ and $A$ are the rotor solidity and rotor disc area, respectively; $d_0$ is the fuselage drag ratio; $\rho$ is the air density; and $v_{xy}(t)$ is the horizontal speed at time $t$, namely $v_{xy}(t) = \sqrt{v_x^2(t) + v_y^2(t)}$. 
More specifically, $P_0=\frac{\delta}{8}\rho sA\Omega^3\zeta^3$ and $P_1=(1+l)\frac{W^{3/2}}{\sqrt{2\rho A}}$,
where $\delta$ is the profile drag coefficient, $\Omega$ is the blade angular velocity in radians/second, $\zeta$ is rotor radius in meter, $l$ is the incremental correction factor to induced power, and $W$ is the aircraft weight in newton. Using \eqref{energy_GPECM}, the UAV traveling energy consumption in joules ($J$) over the whole trajectory is given by
\begin{equation}\label{eq:E_fly}
    {E_{{\mathop{\rm fly}\nolimits} }} = \int_0^{{T}} {{P_{{\rm{fly}}}}} \left( {{\bf{v}}(t)} \right)dt.
\end{equation}

\begin{figure}[t!]
  \centering
  \subfigure[UAV traveling power with different values of vertical speed $|v_z|$.]{
    \includegraphics[width=0.46\linewidth]{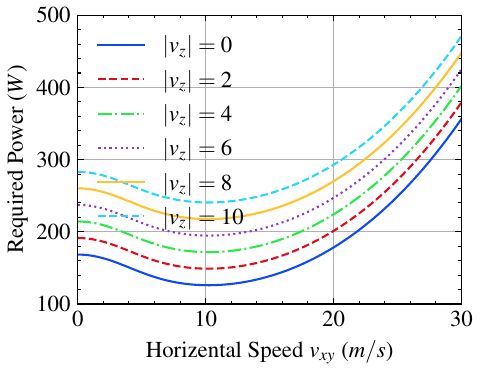}
    \label{fig:velocity_vertical_power}
  }
  \subfigure[UAV traveling power with different values of acceleration $\|\mathbf{a}\|$ when $|v_z|=0$.]{
    \includegraphics[width=0.46\linewidth]{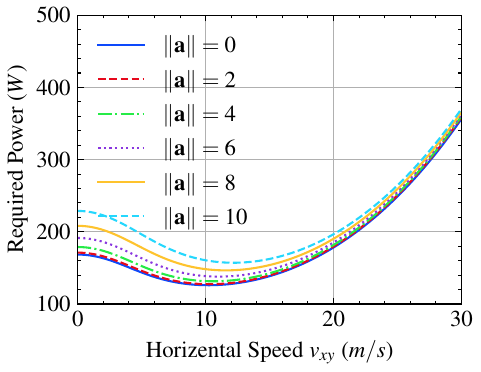}
    \label{fig:velocity_acceleration_power}
  }
  \caption{UAV traveling power study with different vertical speeds and acceleration.}
  \label{fig:model_comparison}
\end{figure}

Fig.~\ref{fig:velocity_vertical_power} illustrates the UAV traveling power with respect to the horizontal speed $v_{xy}$ when the vertical speed $|v_z|$ is given. As seen in Fig. \ref{fig:velocity_vertical_power}, the power consumption during UAV flight is significantly influenced by both its horizontal and vertical speeds \cite{mei20223d}. It is also evident that UAV acceleration impacts power consumption, as discussed in \cite{dai2022energy}, which accounts for the additional power consumed during UAV acceleration. Fig.~\ref{fig:velocity_acceleration_power} illustrates the UAV power consumption model in \cite{dai2022energy} with respect to the horizontal speed $v_{xy}$ when the acceleration $\|\mathbf{a}\|$ is given. As seen in the figure, the impact of speed on power consumption is more pronounced when $\|\mathbf{a}\|$ is small. For the sake of simplicity and to facilitate a clearer analysis, this study does not consider the additional power consumed due to UAV acceleration. However, it is important to note that the proposed MDE-CH in Section \ref{sec:proposed} is universally applicable for various power consumption models, including those in \cite{mei20223d, dai2022energy}.

\subsection{Channel Model for Communication Task }\label{sec:channel_model}

\subsubsection{Air-to-ground channel model}
%Our proposed model uses LoS and non-LoS (NLoS) links. 
The channel model between the UAV and GNs follows the air-to-ground model in \cite{7037248}.
%Let $\mathbf{s}_k = {[s_{k,x}, s_{k,y}, s_{k,z}]}^T\in\mathbb{R}^{3\times 1}$ be $k$-th GN's location, where $k = 1,2,3,...,K$. 
The channel coefficient between two communication points (i.e., $\mathbf{q}(t)$ and $\mathbf{u}_k$) at time $t$ is represented as 
\begin{equation}
{h_k}(t) = \sqrt {{h_{k,{\rm{LS}}}}(t)} {h_{k,{\rm{SS}}}}(t),
\end{equation}
where ${{h_{k,{\rm{LS}}}}(t)}$ and ${h_{k,{\rm{SS}}}}(t)$ are the large-scale and the small-scale fading coefficients, respectively. The small-scale fading coefficient is expressed as a complex random variable with the expectation $\mathbb{E}\big[{{{\left| {{h_{k,{\rm{SS}}}}(t)} \right|}^2}}\big] = 1$. The large-scale fading coefficient is expressed as LoS and non-LoS (NLoS) components respectively, as follows:
\begin{align}\label{path_loss}
    {{h_{k,{\rm{LS}}}}(t)} = \begin{cases}
        \beta_{0}d_k^{-{\alpha}}(t)\text{ for the LoS case},\\
        \kappa \beta_{0}d_k^{-{\alpha}}(t) \text{ for the NLoS case},
    \end{cases}
\end{align}
where $d_k(t) = {\left\|\mathbf{q}(t)-\mathbf{u}_k\right\|}$ is the distance between the UAV and GN $k$ at time $t$, $\beta_{0}>0$ represents the channel power according to the 1-meter reference distance, ${\alpha}\geq 2$ is the path loss exponent, and $\kappa\in(0,1)$ is the additional attenuation factor for the NLoS channel. With given wavelength $\lambda$, $\beta_{0} = {\left(\frac{\lambda}{4\pi}\right)}^2$. 

According to \cite{al2014optimal}, the LoS probability  of GN $k$ is computed as 
\begin{equation}
    {\mathbb{P}_{k,{\rm{LoS}}}}\left( t \right) = \frac{1}{1+ a\operatorname{exp}\left(-b\left({\theta}_k(t) - a\right)\right)},
\end{equation}
where $a>0$ and $b>0$ are the parameters determined by the communication environment given in \cite{7037248},  and ${\theta}_k(t)$ is the elevation angle that is given by
\begin{equation}\label{eq:elevation_angle}
    {\theta}_k(t) = \frac{180}{\pi} \arcsin{\left(\frac{q_z(t)-u_{k,z}}{d_k(t)}\right)},
\end{equation}
where $\arcsin \left(  \cdot  \right)$ denotes the reverse sine function. 
Obviously, the NLoS probability of GN $k$ is given as ${\mathbb{P}_{k,{\rm{NLoS}}}}\left( t \right) = 1 - {\mathbb{P}_{k,{\rm{LoS}}}}\left( t \right)$. Then, the instantaneous achievable data rate in bits per second (bps) from GN $k$ to the UAV is expressed as 
\begin{equation}\label{rate_version1}
    R_k(t) = B \log_2{\left(1 + \frac{{P_{{\rm{tx}}}} {|h_k(t)|}^2}{\sigma^2 \Gamma_0}\right)},
\end{equation}
where $B$ is the bandwidth, $\sigma^2$ is the noise variance, and $\Gamma_0 > 1$ is the channel capacity loss due to the modulation.

\subsubsection{Data requirement for GNs}
Note that $R_k(t)$ in \eqref{rate_version1} contains randomness due to small-scale fading and LoS or NLoS probabilities. Obviously, it is impossible to guarantee a certain amount of data received from each of GNs by adjusting the trajectory of the UAV. To overcome such difficulty, we introduce an approximate expected data rate, which can be determined by the position of the UAV. 
Specifically, the expected channel gain of GN $k$ is given by
\begin{align}
    \mathbb{E}\left[{|h_{k}(t)|}^2\right] &= {\mathbb{P}_{k,{\rm{LoS}}}}\left( t \right)\beta_{0}d_{k}^{-{\alpha}}(t)\nonumber\\ 
    &{~~~}+ \left(1 - \mathbb{P}_{k,\operatorname{LoS}}(t)\right) \kappa \beta_{0} d_{k}^{-{\alpha}}(t)\nonumber\\
    & = {{{{\hat {\mathbb{P}}}}}_{k,{\mathop{\rm LoS}\nolimits} }}(t){\beta _0}d_k^{ - \alpha }(t),
\end{align}
where ${\hat {\mathbb{P}}_{k,{\mathop{\rm LoS}\nolimits} }}(t) = \kappa + (1-\kappa){{ {\mathbb{P}}}_{k,{\rm{LoS}}}}\left( t \right)$ is the regularized LoS probability.
%Let ${P_{{\rm{tx}}}}$ be the transmit power at each GN. 
Consequently, according to \cite{zeng2019energy, goldsmith2005wireless}, we have
\begin{align}\label{mean_data_rate}
    \mathbb{E}\left[R_k(t)\right]&\leq B\log_2\left(1+ \frac{{P_{{\rm{tx}}}} \mathbb{E}\left[{\left|h_k(t)\right|}^2\right]}{\sigma^2 \Gamma_0}\right)\nonumber\\
    &= B\log_2\left(1+ \frac{\gamma_0 {{{\hat {\mathbb{P}}}}_{k,{\mathop{\rm LoS}\nolimits} }}(t)}{{{\left\|\mathbf{q}(t) - \mathbf{u}_k\right\|}^{{\alpha}}}}\right)\nonumber\\
    &:=\tilde{R}_k(t),
\end{align}
where $\gamma_0 = \frac{{P_{{\rm{tx}}}} \beta_0}{\sigma^2 \Gamma_0}$.
From now on, we will use $\tilde{R}_k(t)$ to estimate the achievable data rate at time $t$. Let $R_k^{{\rm{th}}}>0$ be a threshold representing the minimum data rate required for successful communication between the UAV and GN $k$. This threshold is also beneficial for conserving communication energy by avoiding communication initiation for low data rates.
Then the total amount of data in bits received from GN $k$ is given by
\begin{align}
    Q_k &= \int_{0}^{T} \tilde{R}_k(t)\mathbf{1}_{(\tilde{R}_k(t)\geq R_k^{{\rm{th}}})} dt.\label{throuhgput_ver1}
\end{align}

Let us denote the data requirement for GN $k$ by $Q_k^{{\rm{th}}}$. That is, the UAV trajectory $\{\mathbf{q}(t)\}_{t\in[0,T]}$ should satisfy $Q_k\geq Q_k^{{\rm{th}}}$ for all $k\in[1:K]$.

Finally, the total energy consumption in joules ($J$) for completing the communication task is given by
\begin{equation} \label{eq:E_com}
    {E_{{\mathop{\rm com}\nolimits} }} = P_{{\mathop{\rm com}}}\sum\limits_{k = 1}^K\int_{0}^{T}  \mathbf{1}_{(\tilde{R}_k(t)\geq R_k^{{\rm{th}}})}  dt,
\end{equation}
where $P_{{\mathop{\rm com}}}$ denotes the total power consumption during the communication period, including the transmit power $P_{{\rm{tx}}}$ for each GN and processing power at the UAV for data reception \cite{zeng2019energy, arnold2010power}. Here, $\mathbf{1}_{(\cdot)}$ denotes the indicator function, defined by
\begin{equation}
    \mathbf{1}_{(\theta_1\geq \theta_2)}= \begin{cases}
        1 \mbox{ if } \theta_1 \geq \theta_2,\\
        0 \mbox{ otherwise}.
    \end{cases}
\end{equation}

\subsection{UAV Trajectory Optimization for Energy Minimization}
The objective of this work is to minimize UAV energy consumption under various constraints such as UAV physical limitation and communication requirements through joint UAV trajectory and task completion time optimization. To achieve such an objective, we formulate the following optimization problem:
\begin{align}\label{optimization_problem}
    \min_{\{\mathbf{q}(t)\}_{t\in[0,T]}} \left\{E_{\operatorname{fly}} + E_{\operatorname{com}}\right\},
\end{align}
subject to
\begin{align} \label{eq:obtacle_dection}
&\mbox{C1: }\mathbf{q}(0)=\mathbf{q}_{\operatorname{str}}, \nonumber\\
    &\mbox{C2: }\mathbf{q}(T)=\mathbf{q}_{\operatorname{end}},\nonumber\\
    &\mbox{C3: }[q_x(t),q_y(t)]\in \mathcal{U} {~~}\forall t\in(0,T),\nonumber\\
    &\mbox{C4: }{q_z(t)}\in[f_z(q_x(t),q_y(t))+\Delta_z, U_z] {~~}\forall t\in(0,T),\nonumber\\
    &\mbox{{C5}: }v(t) \leq v_{\operatorname{max}} {~~}\forall t\in(0,T),\nonumber\\
    &\mbox{{C6}: }|a_x(t)| \leq a_{x,\operatorname{max}}{~~}\forall t\in(0,T),\nonumber\\
    &\mbox{{C7}: }|a_y(t)| \leq a_{y,\operatorname{max}}{~~}\forall t\in(0,T),\nonumber\\
    &\mbox{{C8}: }|a_z(t)| \leq a_{z,\operatorname{max}}{~~}\forall t\in(0,T),\nonumber\\
    &\mbox{C9: }Q_k \geq Q_k^{{\rm{th}}} {~~}\forall k\in[1:K], \nonumber\\
    &\mbox{C10: } 0 \leq T \leq T_{\operatorname{max}}.
\end{align}
Here, $v_{\operatorname{max}}$  denotes the maximum speed of the UAV and $a_{i,\operatorname{max}}$, $i \in \left\{ {x,y,z} \right\}$ denotes the maximum acceleration along the $(x, y, z)$-axis of the UAV.
The reason for imposing a stricter constraint on acceleration than on speed, i.e., limiting acceleration along the $(x, y, z)$-axis, is to ensure smooth flying speed. Strictly limiting acceleration prevents significant variations or discontinuities in the speed curve, making real-world deployment more manageable. As seen in Fig.~\ref{fig:velocity_acceleration_power}, acceleration has a significant impact on the UAV power consumption, especially at low speeds.

\section{Optimization for Continuous UAV Trajectory}\label{sec:Optimization_for_Continuous_UAV_Trajectory}

In principle, optimizing a continuous UAV trajectory in \eqref{optimization_problem} is impossible. Instead of optimizing $\{\mathbf{q}(t)\}_{t\in[0,T]}$ directly, we first establish a set of discretized control points and generate the corresponding continuous trajectory based on these control points. We then optimize these control points to obtain an optimized $\{\mathbf{q}(t)\}_{t\in[0,T]}$.

\subsection{Trajectory Representation Based on the B\'{e}zier Curve}
In order to generate a continuous UAV trajectory $\{\mathbf{q}(t)\}_{t\in[0,T]}$, we adopt the B\'{e}zier curve interpolation introduced in \cite{song2021improved} \cite{chen2022trajectory}, which is a parametric curve widely used in computer graphics, vehicle streamlines generation, and other fields. %We can easily obtain a smooth curve using the B\'{e}zier curve function as long as control points are determined. 
Firstly, define a set of $M$ control points as
\begin{equation} \label{eq:bezier_curve}
    {\bf{P}} = \left[ {{{\bf{p}}_1},{{\bf{p}}_2}, \ldots ,{{\bf{p}}_M}} \right],
\end{equation}
where $\mathbf{p}_i = [p_{i,x}, p_{i,y}, p_{i,z}]$ for $i\in[1:M]$.
Under given control points $\mathbf{P}$, the B\'{e}zier curve is generated by using the following equation: 
\begin{align} \label{eq:bu}
    \mathbf{b}(u) & = {\left[ {b_x(u),b_y(u),b_z(u)} \right]}, \nonumber\\
    & = \sum_{i=1}^M {\binom{i}{M}{(1-u)}^{M-i}u^i \mathbf{p}_i},
\end{align}
where $u \in \left[ {0,1} \right]$ is an independent variable. 
%$\binom{i}{M}$ is the combination of $i$ and $M$, typically used in statistics to represent all possible ways of selecting $i$ elements from a set of $M$ distinct elements. 
The B\'{e}zier curve has a significant characteristic: it does not generally pass through the middle points ($\mathbf{p}_2$ to $\mathbf{p}_{M-1}$) but always includes the start point $\mathbf{p}_1$ and the end point $\mathbf{p}_M$. This characteristic aids in trajectory design because it ensures controllability over the start and end points of the trajectory.
\begin{figure}[t!]
    \centering
    \includegraphics[width=0.7\linewidth]{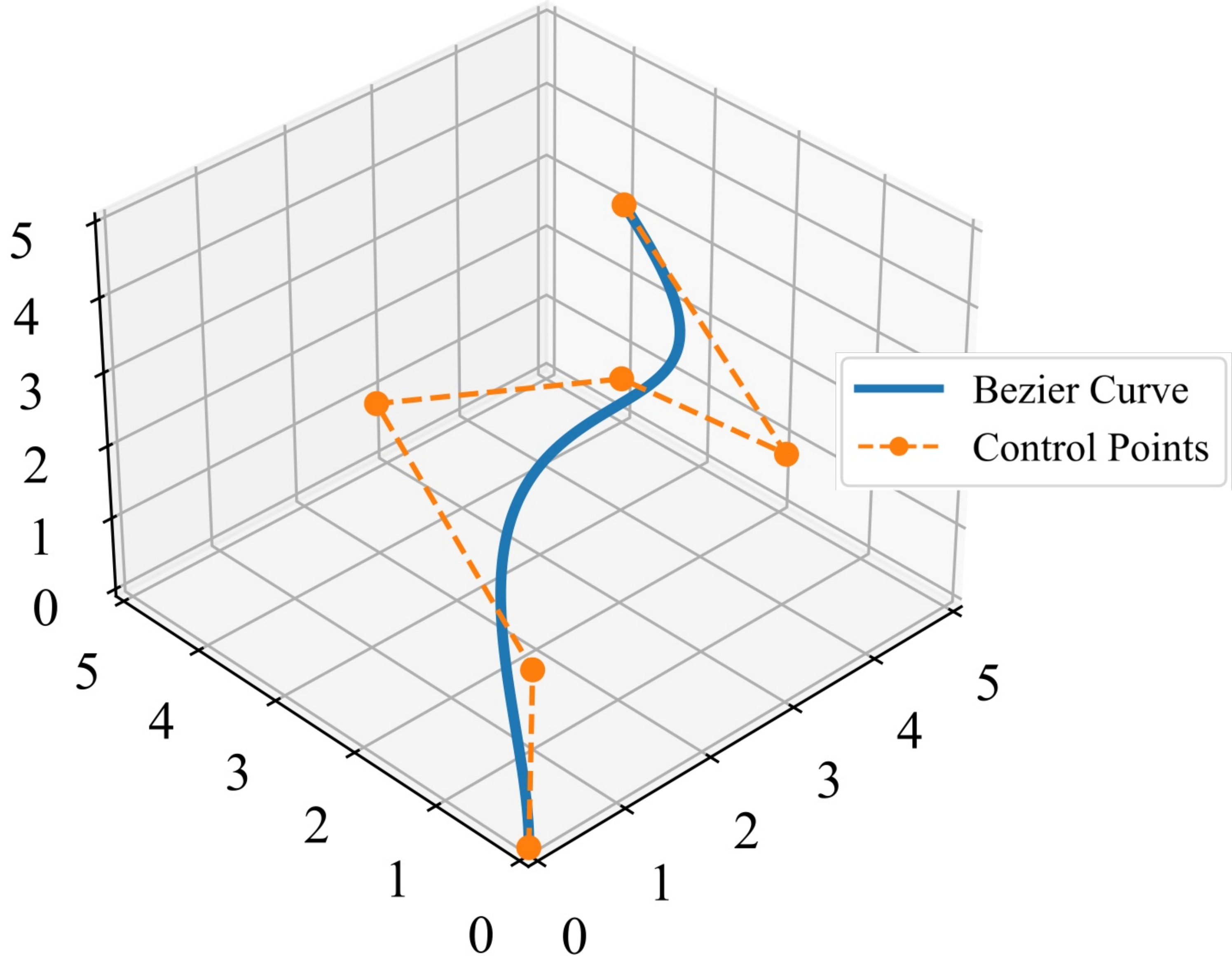}
    \caption{Illustration of an exemplary B\'{e}zier curve with $6$ control points $\mathbf{p}_{1}={[0,0,0]}$, $\mathbf{p}_{2}={[1,1,1]}$, where $\mathbf{p}_{3}={[2,4,2]}$, $\mathbf{p}_{4}={[3,2,3]}$, $\mathbf{p}_{5}={[4,1,2]}$ and $\mathbf{p}_{6}={[5,4,3]}$.}
    \label{fig:bezier_demo}
\end{figure}
Fig.~\ref{fig:bezier_demo} illustrates an exemplary trajectory of the B\'{e}zier curve generated by six control points. 

To apply the B\'{e}zier curve interpolation in \eqref{eq:bezier_curve} and \eqref{eq:bu}, we introduce a normalized time $\bar t = t/{{T}}\in[0,1]$. Then the relationship between $\mathbf{b}(\bar{t})$ and $\mathbf{q}(t)$ is given by
\begin{equation}\label{eq:b2q}
    \mathbf{b}(\bar{t}) = \mathbf{q}(t).
\end{equation}
Furthermore, we consider a set of discretized time indices $\bar{\mathbf{t}}=[0,\Delta_{\bar{t}},2\Delta_{\bar{t}},\ldots, 1]$ to evaluate $E_{\operatorname{fly}}$ and $E_{\operatorname{com}}$ in \eqref{optimization_problem} and the constraints in \eqref{eq:obtacle_dection} based on $\bar{t}\in \bar{\mathbf{t}}$, where $\Delta_{\bar t} = \Delta_t/T$ and the cardinality of $\bar{\mathbf{t}}$ is given by $n = T/\Delta_t + 1$ for a given $\Delta_t>0$.
Specifically, from \eqref{eq:v_deri} to \eqref{eq:a}, we have
\begin{align}
    \bar{\mathbf{v}}(\bar{t}) &= {\left[\bar{v}_{x}(\bar{t}), \bar{v}_{y}(\bar{t}), \bar{v}_{z}(\bar{t})\right]} = \frac{\mathbf{b}(\bar{t}+\Delta_{\bar{t}})-\mathbf{b}(\bar{t})}{ T\Delta_{\bar{t}}}, \label{eq:bezier_vec_v}\\
    \bar{v}(\bar{t}) &= \|\bar{\mathbf{v}}(\bar{t})\|, \label{eq:bezier_v}\\
    \bar{\mathbf{a}}(\bar{t})  &= {\left[\bar{a}_{x}(\bar{t}), \bar{a}_{y}(\bar{t}), \bar{a}_{z}(\bar{t})\right]}= \frac{\bar{\mathbf{v}}(\bar{t}+\Delta_{\bar{t}}) -\bar{\mathbf{v}}(\bar{t})}{T\Delta_{\bar{t}}}\label{eq:bezier_a}
\end{align}
for $\bar{t}\in \bar{\mathbf{t}}$.
Also, from \eqref{eq:E_fly}, \eqref{throuhgput_ver1}, and \eqref{eq:E_com}, we have

\begin{align}
    {\bar{E}_{{\mathop{\rm fly}\nolimits} }} &= T\Delta_{\bar{t}}\sum_{\bar{t}\in \bar{\mathbf{t}}}{{P_{{\rm{fly}}}}} \left( {{\bf{v}}(\bar{t})} \right), \\
    \bar{Q}_k &= T\Delta_{\bar{t}} \sum_{\bar{t}\in \bar{\mathbf{t}}} \tilde{R}_k(\bar{t})\mathbf{1}_{(\tilde{R}_k(\bar{t})\geq R_k^{{\rm{th}}})}, \label{eq:Q_k_bar}\\
    {\bar{E}_{{\mathop{\rm com}\nolimits} }} &= T\Delta_{\bar{t}}P_{{\mathop{\rm com}}}\sum\limits_{k = 1}^K\sum_{\bar{t}\in \bar{\mathbf{t}}} \mathbf{1}_{(\tilde{R}_k(\bar{t})\geq R_k^{{\rm{th}}})}.
\end{align}

%Compared with the discretized UAV trajectory as in the existing works in Table~\ref{tab:comparison_work}, considering continuous UAV trajectory based on the B\'{e}zier curve can decrease the number of decision variables. More specifically, under the consideration of the discretized UAV trajectory, the UAV trajectory are directly denoted by discretized points $\mathbf{q}(t_i)\in \mathbb{R}^{3}$, $i=1,2,\cdots$, which are decision variables. Further, to achieve more smooth UAV trajectory, the number of decision variables needs to increase. Whereas, the advantage of using B\'{e}zier curve is that its smooth degree depends on the interpolation number rather than control points (decision variables).

It is worthwhile to mention that the number of control points $M$ can be much smaller than the cardinality of $\bar{\mathbf{t}}$, i.e., $T/\Delta_t + 1$.
Hence the proposed trajectory representation based on the B\'{e}zier curve can efficiently optimize a continuous UAV trajectory.

\subsection{Revised Optimization Problem} \label{subsec:revised_opt}
Since we generate a UAV trajectory through the B\'{e}zier curve, the control variables $\{\mathbf{q}(t)\}_{t\in[0,T]}$ for the original optimization problem \eqref{optimization_problem} and \eqref{eq:obtacle_dection} should be replaced with the control points of the B\'{e}zier curve, i.e., ${\bf{P}} = [{{\bf{p}}_1},{{\bf{p}}_2}, \cdots ,{{\bf{p}}_{M}}]$. As a consequence, \eqref{optimization_problem} and \eqref{eq:obtacle_dection} can be rewritten as
\begin{align}\label{optimization_problem_re}
    \mathop {\min }\limits_{T,{\bf{P}} = [{{\bf{p}}_1},{{\bf{p}}_2}, \cdots ,{{\bf{p}}_{M }}]} \{{\bar{E}_{{\mathop{\rm fly}\nolimits} }} + {\bar{E}_{{\mathop{\rm com}\nolimits} }}\}
\end{align}
subject to
\begin{align}\label{eq:constraint_re}
&\mbox{C1: }\mathbf{b}(0)=\mathbf{q}_{\operatorname{str}}, \nonumber\\
    &\mbox{C2: }\mathbf{b}(1)=\mathbf{q}_{\operatorname{end}},\nonumber\\
    &\mbox{C3: }[b_x({\bar t}),b_y({\bar t})]\in \mathcal{U} {~~}\forall \bar{t}\in \bar{\mathbf{t}},\nonumber\\
    &\mbox{C4: }{b_z({\bar t})}\in[f_z(b_x({\bar t}),b_y({\bar t}))+\Delta_z, U_z] {~~}\forall \bar{t}\in \bar{\mathbf{t}},\nonumber\\
    &\mbox{{C5}: }\bar{v}({\bar t}) \leq v_{\operatorname{max}} {~~}\forall \bar{t}\in \bar{\mathbf{t}},\nonumber\\
    &\mbox{{C6}: }|\bar{a}_x({\bar t})| \leq a_{x,\operatorname{max}}{~~}\forall \bar{t}\in \bar{\mathbf{t}},\nonumber\\
    &\mbox{{C7}: }|\bar{a}_y({\bar t})| \leq a_{y,\operatorname{max}}{~~}\forall \bar{t}\in \bar{\mathbf{t}},\nonumber\\
    &\mbox{{C8}: }|\bar{a}_z({\bar t})| \leq a_{z,\operatorname{max}}{~~}\forall \bar{t}\in \bar{\mathbf{t}},\nonumber\\
    &\mbox{C9: }\bar{Q}_k \geq Q_k^{{\rm{th}}} {~~}\forall k\in[1:K], \nonumber\\
    &\mbox{C10: } 0\leq T \leq T_{\operatorname{max}},
\end{align}
where the relation between $\mathbf{P}$ and $\mathbf{b}(u)$ is given by \eqref{eq:bu}.
\section{Optimization Via Matrix-based DE with Constraint Handle} \label{sec:proposed}
% [non-convex 문제이고 no-hard 문제이다 --> global optimization technqiue으로 matrix-bases DE technqiue을 쓰겠다]
The optimization problem presented in \eqref{optimization_problem_re} and \eqref{eq:constraint_re} is evidently non-convex, and obtaining its optimal solution in real-time processes may be unattainable. To address this challenge, various global optimization techniques can be considered. Among the diverse methods highlighted in \cite{stork2020new}, we adopt an evolutionary computation-based optimization, a popular and promising approach within this field.
Inspired by \cite{zhan2021matrix}, which introduces the matrix-based particle swarm optimizer (MPSO) and matrix-based genetic algorithm (MGA), we propose a matrix-based differential evolution (MDE) with constraint handle (CH), called MDE-CH, to optimize the model presented in this paper. The proposed MDE-CH algorithm is an extended version of the DE algorithm \cite{6621004}. The first extension involves a matrix-based representation of candidate solutions and their parameters that is able to offer the advantage of parallelizing and synchronizing multiple solutions within a matrix. 
%\textcolor{red}{This capability facilitates efficient exploration of the search space and swift identification of promising regions, ultimately leading to faster convergence in various optimization problems.} 
The second extension is to provide an efficient method of handling different types of constraints under such matrix-based operations required for the constrained optimization in this paper.  

\begin{table*}[ht!]
\centering
\caption{Matrix operations used for the proposed MDE-CH algorithm.}\label{tab: matrix_index}
\setlength{\tabcolsep}{6mm}{
\begin{tabular}{cl}
\toprule
Operation & Description    \\ \midrule
$\mathbf{A}_{[i, j]}$ & The element in the $i$th row and the $j$th column of $\mathbf{A}$.\\

$\mathbf{A}_{[i, \cdot]}$  & $ \mathbf{A}_{[i, \cdot]}=\left[\mathbf{A}_{[i, 1]}, \mathbf{A}_{[i, 2]},\cdots, \mathbf{A}_{[i, c_1]}\right]$.  \\

$\mathbf{A}_{[\cdot, j]}$  &  $\mathbf{A}_{[\cdot, j]} = {\left[\mathbf{A}_{[1, j]}, \mathbf{A}_{[2, j]},\cdots, \mathbf{A}_{[c_2, j]}\right]}^T$.\\

$\mathbf{A}_{[\mathbf{i}, \cdot]}$& For $\mathbf{i}\in {\{1, 2,\cdots, c_1\}}^{\rho_1}$, $\mathbf{A}_{[\mathbf{i}, \cdot]} = {\left[\mathbf{A}^T_{[i_1, \cdot]},\mathbf{A}^T_{[i_2, \cdot]},\cdots, \mathbf{A}^T_{[i_{\rho_1}, \cdot]}\right]}^{T}$, where $\mathbf{i} = {[i_1, i_2, \cdots, i_{\rho_1}]}$.\\

$\mathbf{A}_{[\cdot, \mathbf{j}]}$ &For $\mathbf{j}\in {\{1, 2,\cdots, c_2\}}^{\rho_2}$, $\mathbf{A}_{[\cdot, \mathbf{j}]} = {\left[\mathbf{A}_{[\cdot, j_1]},\mathbf{A}_{[\cdot, j_2]},\cdots, \mathbf{A}_{[\cdot, j_{\rho_2}]}\right]}$, where $\mathbf{j} = {[j_1, j_2, \cdots, j_{\rho_2}]}$.\\

$\mathbf{A}_{[\mathbf{i},\mathbf{j}]}$ &For $\mathbf{i}\in {\{1, 2,\cdots, c_1\}}^{\rho}$ and $\mathbf{j}\in {\{1, 2,\cdots, c_2\}}^{\rho}$, $\mathbf{A}_{[\mathbf{i},\mathbf{j}]} = {\left[\mathbf{A}_{[i_1, j_1]},\mathbf{A}_{[i_2, j_2]},\cdots, \mathbf{A}_{[i_{\rho}, j_\rho]}\right]}$,\\

&where $\mathbf{i} = {[i_1, i_2, \cdots, i_{\rho}]}$ and $\mathbf{j} = {[j_1, j_2, \cdots, j_{\rho}]}$.\\

\bottomrule
\end{tabular}
}
\end{table*}

In the following, we explain the detailed description of the proposed MDE-CH algorithm with its four main operations: initialization, mutation, crossover, and selection. Then, we provide how to handle constraints to be satisfied during these matrix-based operations. We introduce the notations used for vector and matrix operations. For a vector $\mathbf{a}$, its $i$th element is denoted by $\mathbf{a}_{[i]}$. 
The element-wise multiplication of two equal-sized matrices is denoted by $\circ$. The $c_1\times c_2$ all-ones matrix and all-zeros matrix are denoted by $\mathbf{J}_{c_1\times c_2}$ and $\mathbf{0}_{c_1\times c_2}$, respectively. The transpose operator is given by ${(\cdot)}^T$. We employ $(\cdot)\overset{*}{<}(\cdot)$ to perform the element-wise comparison of two equal-sized matrices and yield a Boolean indicator matrix. Specifically, each element in the Boolean matrix takes the value of one if the corresponding element in the left-hand side matrix is smaller than that of the right-hand side matrix; otherwise, it takes the value of zero. In the same manner, we define $(\cdot)\overset{*}{>}(\cdot)$ and $(\cdot)\overset{*}{=}(\cdot)$.
For the $c_1\times c_2$ Boolean matrix  $\mathbf{A}$, its reverse Boolean matrix is denoted by $\overline{\mathbf{A}} = \mathbf{J}_{c_1\times c_2} - \mathbf{A}$. The matrix index operations are given in Table \ref{tab: matrix_index}. Besides, we use $\mathcal{U}(L_1, L_2)$ to represent the uniform distribution ranging from $L_1$ to $L_2$. %We regulate that if the variable or parameters with a shape similar to $\mathbb{R}^{(\cdot)}$, we call them vector, and if the shape with similar to $\mathbb{R}^{(\cdot)\times (\cdot)}$, we call them matrix.

%\subsection{Discretization}\label{subsec:discretization}
%The discretization of our model is easy to implement. We let $\Delta t = \frac{T}{n-1}$, where $n-1$ is the segment number. Thus, we can get a sequence of $t_i$ ($i = 0,1,\cdots, n$) from 0 to $T$, i.e., $t_i\in {[0:T]}^T$.
%The trajectory $\mathbf{q}(t)$ can be express as a matrix $\mathbf{q}\in\mathbb{R}^{3 \times n}$ calculated by $\mathbf{q}(t_i)$. The velocity matrix is $\mathbf{v}\in\mathbb{R}^{3\times (n-1)}$, and the acceleration matrix is $\mathbf{a}\in\mathbb{R}^{3\times (n-2)}$.
\subsection{Matrix-Based DE}\label{sec:MDE}
\emph{\textbf{Population Representation:}}
%The core of the EC algorithm is the population that all evolutionary operators act on. 
In MDE, the population is represented by a matrix including optimization parameters. 
Specifically, the matrix-based population can be given by $\mathbf{X}^{g}\in\mathbb{R}^{N\times D}$ in which each row vector represents an individual and each element in columns corresponds to an optimization parameter, where $g$ is the generation number, $N$ is the number of individuals, and $D$ is the dimension size. 
Let $\mathbf{U}\in \mathbb{R}^{1\times D}$ and $\mathbf{L}\in \mathbb{R}^{1\times D}$ be vectors representing the upper and lower bounds of each dimension, respectively.

For the considered UAV trajectory optimization in Section~\ref{subsec:revised_opt}, the population matrix is presented as 
\begin{equation}\label{eq:pop_mat}
    \mathbf{X}^{g} = \begin{bmatrix}
        \mathbf{p}^{g}_{1,2} & \mathbf{p}^{g}_{1,3} & \cdots & \mathbf{p}^{g}_{1,M-1} & T^{g}_1\\
        \mathbf{p}^{g}_{2,2} & \mathbf{p}^{g}_{2,3} & \cdots & \mathbf{p}^{g}_{2,M-1} & T^{g}_2\\
        \vdots & \vdots & \ddots & \vdots & \vdots\\
        \mathbf{p}^{g}_{N,2} & \mathbf{p}^{g}_{N,3} & \cdots & \mathbf{p}^{g}_{N,M-1} & T^{g}_N
    \end{bmatrix},
\end{equation}
where $D = 3(M - 2) + 1$. Here, each row vector consists of a set of optimization parameters in \eqref{optimization_problem_re} except $\mathbf{p}_1$ and $\mathbf{p}_M$, which is set as $\mathbf{q}_{\operatorname{str}}$ and $\mathbf{q}_{\operatorname{end}}$ respectively to satisfy C1 and C2 in \eqref{eq:constraint_re}. 
Moreover, $\mathbf{U}$ and $\mathbf{L}$ are expressed as 
\begin{equation} \label{eq: upper_lower_vector}
\begin{aligned} 
    \mathbf{U} &= [U_x, U_y, U_z, U_x, U_y, U_z, \cdots, U_x, U_y, U_z, T_{\operatorname{max}}], \\
    \mathbf{L} &= [0, 0, 0, 0,0,0,\cdots,0,0,0, 0].
\end{aligned} 
\end{equation}

\emph{\textbf{Initialization:}} Let $\mathbf{R}_{N \times D}$ be the $N\times D$ random matrix whose elements follow $\mathcal{U}(0,1)$.
Then the population matrix of the first generation is constructed as 
\begin{equation}\label{eq:ini_pop}
    \mathbf{X}^{1} = \left[\mathbf{J}_{N\times 1}(\mathbf{U} - \mathbf{L})\right]  \circ \mathbf{R}_{N\times D}
    + \mathbf{J}_{N\times 1}\mathbf{L}.
\end{equation}
Note that the matrix $\mathbf{J}_{N\times 1}$ facilitates the reshaping of $\mathbf{U}$ and $\mathbf{L}$ from dimensions $1\times D$ to $N\times D$.
After initialization, the elements in $\mathbf{X}^1$ span the range between the upper and lower bounds for each dimension.

\emph{\textbf{Mutation:}} The aim of the mutation operation is to generate new individuals for exploring uncharted searching areas. Here, we primarily introduce the \textit{DE/rand/1} version. In the original DE, the main parameter is the amplification factor $\lambda$, typically ranging from $0$ to $2$. In MDE, $\lambda$ should be a matrix derived by 
\begin{equation}\label{eq:amp_fac}
    \mathbf{\Lambda} = \lambda \mathbf{J}_{N\times D}.
\end{equation}
Then, the donor matrix is generated by
\begin{equation} \label{eq: Y_g}
    \mathbf{Y}^{g} = \mathbf{X}^{g}_{[\mathbf{r}_{1}, \cdot]} + \mathbf{\Lambda} \circ \left(\mathbf{X}^{g}_{[\mathbf{r}_{2}, \cdot]} - \mathbf{X}^{g}_{[\mathbf{r}_{3}, \cdot]}\right).
\end{equation}
Here, $\mathbf{r}_{i}\in\{1,2,3,\cdots, N\}^{N}$ is the random permutation vector to shuffle the set of $N$ individuals (row vectors in $\mathbf{X}^{g}$) based on the order in $\mathbf{r}_{i}$ for $i=1,2,3$.

Due to addition and subtraction operations in \eqref{eq: Y_g}, some elements in $\mathbf{Y}^{g}$ might exceed their upper bounds or fall below their lower bounds, rendering these individuals as infeasible solutions. Therefore, it is essential to enforce that each element remains within the specified range of the upper and lower bounds.
For this purpose, let us first construct two Boolean matrices $\mathbf{\Omega}_U$ and $\mathbf{\Omega}_L$ as  
\begin{equation} \label{eq:sigma_upper_lower}
    \begin{aligned}
        \mathbf{\Omega}_U &= \mathbf{Y}^{g} \overset{*}{<} \mathbf{J}_{N\times 1}\mathbf{U},\\
        \mathbf{\Omega}_L &= \mathbf{Y}^{g} \overset{*}{<} \mathbf{J}_{N\times 1}\mathbf{L},
    \end{aligned}
\end{equation}
Note that if the elements in $\mathbf{Y}^{g}$ are less than the values in $\mathbf{J}_{N\times 1}\mathbf{U}$, the corresponding elements in $\mathbf{\Omega}_U$ become ones; otherwise, they become zeros. Similarly, the elements in $\mathbf{\Omega}_L$ are determined.
Then, the correction of the out-of-bound values in $\mathbf{Y}^{g}$ is performed by updating 
\begin{equation}\label{eq: repair_ul}
    \begin{aligned}
        &\mathbf{Y}^{g} \leftarrow \mathbf{Y}^g \circ \mathbf{\Omega}_U + (\mathbf{J}_{N\times 1}\mathbf{U}) \circ \overline{\mathbf{\Omega}}_U, \\
        &\mathbf{Y}^{g} \leftarrow \mathbf{Y}^g \circ \overline{\mathbf{\Omega}}_L + (\mathbf{J}_{N\times 1} \mathbf{L}) \circ \mathbf{\Omega}_L,
    \end{aligned}
\end{equation}
where $\overline{\mathbf{\Omega}}_U$ and $\overline{\mathbf{\Omega}}_L$ are the reverse Boolean matrices of ${\mathbf{\Omega}}_U$ and ${\mathbf{\Omega}}_L$, respectively.
%Till now, we get two population matrices $\mathbf{P}^{g}$ and $\mathbf{P}^{g*}$ in the $g$th generation. 

\emph{\textbf{Crossover:}}
The main parameter for the crossover procedure is the crossover rate $\gamma$, which ranges between $(0,1)$. Similar to the amplification factor $\lambda$ in the mutation procedure, $\gamma$ should be reshaped from a scalar to a matrix to calculate with the two population matrices $\mathbf{X}^{g}$ and $\mathbf{Y}^{g}$, as 
\begin{equation}\label{eq:cro_rate}
    \mathbf{\Gamma} = \gamma \mathbf{J}_{N\times D},
\end{equation}
where $\mathbf{\Gamma}$ is the $N\times D$ matrix with all elements equal to $\gamma$.
Then, to determine which elements in $\mathbf{Y}^{g}$ will be replaced with the corresponding elements in $\mathbf{X}^{g}$, thus generating the final trial matrix $\mathbf{W}^{g}$, we first construct
\begin{equation}
    \mathbf{\Theta} = \mathbf{R}_{N\times D} \overset{*}{<} \mathbf{\Gamma}.
\end{equation}
Furthermore, to ensure that each individual can participate in the crossover procedure, we additionally update some elements in $\mathbf{\Theta}$ as 
\begin{equation}\label{eq:assign_1}
    \mathbf{\Theta}_{[\mathbf{a}_{1}, \mathbf{a}_{2}]} \leftarrow \mathbf{J}_{N\times 1}, 
\end{equation}
where $\mathbf{a}_1 = {[1:N]}^T$ and $\mathbf{a}_2$ is the $N\times 1$ random vector in which each element follows  $\mathcal{U}(1,D)$.
Consequently, the trial population matrix $\mathbf{W}^g$ is derived as
\begin{equation}\label{eq:w_g}
    \mathbf{W}^{g} = \mathbf{Y}^{g} \circ \mathbf{\Theta} + \mathbf{X}^{g} \circ \overline{\mathbf{\Theta}}, 
\end{equation}
where $\overline{\mathbf{\Theta}}$ is the reverse Boolean matrix of $\mathbf{\Theta}$. 

\emph{\textbf{Selection:}}
The original objective function ${\bar{E}_{{\mathop{\rm fly}\nolimits} }} + {\bar{E}_{{\mathop{\rm com}\nolimits} }}$ in \eqref{optimization_problem} is used for the selection procedure. For convenience, the objective function can be re-defined as $\mathbf{f}_e: \mathbb{R}^{N\times D} \rightarrow \mathbb{R}^{N\times 1}$ as the vector-wise function to evaluate the fitness values of $\mathbf{W}^{g}$ and $\mathbf{X}^{g}$ as $\mathbf{f}_e(\mathbf{W}^{g})$ and $\mathbf{f}_e(\mathbf{X}^{g})$. Then, we compare $\mathbf{f}_e(\mathbf{W}^{g})$ and $\mathbf{f}_e(\mathbf{X}^{g})$ to determine which individual can enter into the next generation by
\begin{equation}\label{eq:update_pop}
    \mathbf{X}^{g+1} = \mathbf{W}^{g} \circ \left(\mathbf{\Omega}_E \mathbf{J}_{1\times D}\right) + \mathbf{X}^{g} \circ \left(\overline{\mathbf{\Omega}}_E \mathbf{J}_{1\times D}\right),
\end{equation}
where
\begin{equation}
    \mathbf{\Omega}_E = \mathbf{f}_e(\mathbf{W}^{g}) \overset{*}{<} \mathbf{f}_e(\mathbf{X}^{g})
\end{equation}
 and $\overline{\mathbf{\Omega}}_E$ is the reverse Boolean matrix of $\mathbf{\Omega}_E$.

\subsection{Selection under Constraints} \label{subsec:seleciton_constraints}
Previously, we introduced a general matrix-based selection procedure. In this subsection, we present detailed selection steps that reflect the set of constraints in \eqref{eq:constraint_re}.

\subsubsection{Constraints C1, C2, C3, and C10}
The constraints C1 and C2 are always satisfied if ${\mathbf{p}_1}=\mathbf{q}_{\operatorname{str}}$ and ${\mathbf{p}_M}=\mathbf{q}_{\operatorname{end}}$ because of the characteristics of the B\'{e}zier curve. For C3, and C10, we set the lower and upper and bounds as $(0,U_x)$, $(0,U_y)$, and $(0,T_{\operatorname{max}})$ in \eqref{eq: upper_lower_vector} and then the out-of-bound values are corrected during the mutation procedure as in \eqref{eq:sigma_upper_lower} and \eqref{eq: repair_ul}. Hence, the constraints C3 and C10 are satisfied.

\subsubsection{Constraints C4 to C9}
Let us now explain the remaining constraints. 
For these constraints, we introduce a non-negative violation function and choose the one with a smaller violation value during the selection procedure if both individuals in \eqref{eq:update_pop} are infeasible, rather than considering their objective values. 
For such purpose, denote the violation function of the $i$th constraint by $C_i(\mathbf{x})\in[0,+\infty)$ for $i\in[4:9]$ when an individual $\mathbf{x}$ is given. Then the total violation function is given by 
\begin{equation} \label{eq:total_violation}
    \phi(\mathbf{x}) = \sum_{i=4}^9 \omega_iC_i(\mathbf{x}),
\end{equation}
where $\{\omega_i\}_{i\in[4:9]}$ are weighting coefficients satisfying that $\sum_{i=4}^9 \omega_i =1$ with $\omega_i>0$ for all $i\in[4:9]$. 
%{\color{blue}{Let us further explain how the comparison step works. Two individuals with respect to $\phi(\mathbf{x}_{1})$ and $\phi(\mathbf{x}_{2})$ have totally three comparison results as
%\begin{align}
%	\begin{cases}
%		\phi(\mathbf{x}_{1}) = \phi(\mathbf{x}_{2}),\\
%		\phi(\mathbf{x}_{1}) < \phi(\mathbf{x}_{2}),\\
%		\phi(\mathbf{x}_{1}) > \phi(\mathbf{x}_{2}).
%	\end{cases}
%\end{align}
%The two individuals will be compared for their objective function values if $\phi(\mathbf{x}_{1}) = \phi(\mathbf{x}_{2})$, and the one with the smaller value will be accepted to enter into the next generation. Otherwise, the two individuals are directly compared in their constraint violation degree, and the one with the smaller value will be accepted to enter into the next generation.}}   

Now, we introduce how to calculate each violation function. For C4, the constraint $b_{z}(\bar{t}) \leq U_z$ is satisfied for all $\bar{t}\in \bar{\mathbf{t}}$ because of \eqref{eq:sigma_upper_lower} and \eqref{eq: repair_ul} during the mutation procedure as the same manner applied for C3 and C10. Then, to define the violation function for $ b_{z}(\bar{t}) \geq f_{z}(b_{x}\left(\bar{t}), b_{y}(\bar{t})\right) + \Delta_{z}$, let 
\begin{align}\label{eq:bx_by_bz}
 \begin{cases}
        \mathbf{b}_{x} = [b_x(0), b_x(\Delta_{\bar{t}}), b_x(2\Delta_{\bar{t}}), \dots, b_x((n-1)\Delta_{\bar{t}})],\\
        \mathbf{b}_{y} = [b_y(0), b_y(\Delta_{\bar{t}}), b_y(2\Delta_{\bar{t}}), \dots, b_y((n-1)\Delta_{\bar{t}})],\\
        \mathbf{b}_{z} = [b_z(0), b_z(\Delta_{\bar{t}}), b_z(2\Delta_{\bar{t}}), \dots, b_z((n-1)\Delta_{\bar{t}})],
        \end{cases}
\end{align}
and denote the vector-wise ground altitude function as $\mathbf{f}_z:\mathcal{U}^{1\times n}\rightarrow \mathbb{R}^{1\times n}$.
The violation function of C4 is expressed as 
\begin{multline}
    {C_4} = \Bigl[\left(\left(\mathbf{f}_z(\mathbf{b}_{x}, \mathbf{b}_{y}) + \Delta_z\mathbf{J}_{1\times n} - \mathbf{b}_{z}\right)\overset{*}{>}\mathbf{0}_{1\times n}\right)\\
    \circ\left(\mathbf{f}_z (\mathbf{b}_{x}, \mathbf{b}_{y}) + \Delta_z\mathbf{J}_{1\times n} - \mathbf{b}_{z}\right)\Bigr]\mathbf{J}_{n\times 1}.
\end{multline}

The constraints C5 to C8 are about the UAV's dynamics. For C5, we represent $\mathbf{v}^{*} \in \mathbb{R}^{1 \times (n-1)}$ as the UAV's speed along the trajectory according to \eqref{eq:bezier_v}, which is given by
\begin{equation}
    \mathbf{v}^{*} = [\bar{v}(0), \bar{v}(\Delta_{\bar{t}}), \bar{v}(2\Delta_{\bar{t}}), \dots, \bar{v}((n-2)\Delta_{\bar{t}})].
\end{equation}
Note that the endpoint speed cannot be calculated because the speed calculation uses two adjacent points $\mathbf{b}(\bar{t})$ and $\mathbf{b}(\bar{t} + \Delta_{\bar{t}})$. Hence, the violation function of C5 is expressed as
\begin{multline}
    {C_5} = \Bigl[\left(\left(\mathbf{v}^{*} - v_{\operatorname{max}}\mathbf{J}_{1\times (n-1)}\right) \overset{*}{>}\mathbf{0}_{1\times (n-1)} \right) \\
    \circ \left(\mathbf{v}^{*} - v_{\operatorname{max}}\mathbf{J}_{1\times (n-1)}\right)\Bigr]\mathbf{J}_{(n-1)\times 1}.
\end{multline}

The constraints C6 to C8 are about acceleration.
Let 
\begin{align}
    \begin{cases}
        \mathbf{a}_{x} = [|\bar{a}_x(0)|, |\bar{a}_x(\Delta_{\bar{t}})|, |\bar{a}_x(2\Delta_{\bar{t}})|, \dots, |\bar{a}_x((n-3)\Delta_{\bar{t}})|],\\
        \mathbf{a}_{y} = [|\bar{a}_y(0)|, |\bar{a}_y(\Delta_{\bar{t}})|, |\bar{a}_y(2\Delta_{\bar{t}})|, \dots, |\bar{a}_y((n-3)\Delta_{\bar{t}})|],\\
        \mathbf{a}_{z} = [|\bar{a}_z(0)|, |\bar{a}_z(\Delta_{\bar{t}})|, |\bar{a}_z(2\Delta_{\bar{t}})|, \dots, |\bar{a}_z((n-3)\Delta_{\bar{t}})|].\\
    \end{cases}
\end{align}
Then, the violation function for C6 to C8 are calculated by 
\begin{multline}
    {C_6} = \Bigl[\left(\left(\mathbf{a}_{x} - a_{x,\operatorname{max}}\mathbf{J}_{1\times (n-2)}\right) \overset{*}{>}\mathbf{0}_{1\times (n-2)} \right)\\
    \circ \left(\mathbf{a}_{x} - a_{x,\operatorname{max}}\mathbf{J}_{1\times (n-2)}\right)\Bigr]\mathbf{J}_{(n-2)\times 1},
\end{multline}
\begin{multline}
    {C_7} = \Bigl[\left(\left(\mathbf{a}_{y} - a_{y,\operatorname{max}}\mathbf{J}_{1\times (n-2)}\right) \overset{*}{>}\mathbf{0}_{1\times (n-2)} \right) \\
    \circ \left(\mathbf{a}_{y} - a_{y,\operatorname{max}}\mathbf{J}_{1\times (n-2)}\right)\Bigr]\mathbf{J}_{(n-2)\times 1},
\end{multline}
\begin{multline}
    {C_8} = \Bigl[\left(\left(\mathbf{a}_{z} - a_{z,\operatorname{max}}\mathbf{J}_{1\times (n-2)}\right) \overset{*}{>}\mathbf{0}_{1\times (n-2)} \right)\\
    \circ \left(\mathbf{a}_{z} - a_{z,\operatorname{max}}\mathbf{J}_{1\times (n-2)}\right)\Bigr]\mathbf{J}_{(n-2)\times 1}.
\end{multline}
For C9, denote
\begin{equation}
    \mathbf{Q} = {[\bar{Q}_1, \bar{Q}_2, \bar{Q}_3, \dots, \bar{Q}_K]}
\end{equation}
and 
\begin{equation}
    \mathbf{Q}^{{\rm{th}}} = {[Q_1^{{\rm{th}}}, Q_2^{{\rm{th}}}, Q_3^{{\rm{th}}}, \dots, Q_K^{{\rm{th}}}]}.
\end{equation}
Then, the violation function of C9 is calculated as 
\begin{align}
    {C_9} = \left[\left(\left(\mathbf{Q}^{\mathrm{th}} - \mathbf{Q}\right) \overset{*}{>} \mathbf{0}_{1\times K} \right) \circ \left(\mathbf{Q}^{\mathrm{th}} - \mathbf{Q}\right)\right]\mathbf{J}_{K\times 1}.
\end{align}

\subsubsection{Selection based on the violation function} Let us now explain the modified section procedure incorporating the total violation function in \eqref{eq:total_violation}.
First, define the $N\times 1$ violation vectors for  $\mathbf{X}^g$ and $\mathbf{W}^g$ as 
\begin{align}
        \mathbf{\Phi}_{\mathbf{X}^g} &= {\left[\phi\left(\mathbf{X}^{g}_{[1,\cdot]}\right), \phi\left(\mathbf{X}^{g}_{[2,\cdot]}\right), \dots, \phi\left(\mathbf{X}^{g}_{[N,\cdot]}\right)\right]}^T, \\
        \mathbf{\Phi}_{\mathbf{W}^g} &= {\left[\phi\left(\mathbf{W}^{g}_{[1,\cdot]}\right), \phi\left(\mathbf{W}^{g}_{[2,\cdot]}\right), \dots, \phi\left(\mathbf{W}^{g}_{[N,\cdot]}\right)\right]}^T.
\end{align}
The original objective function values will be used for selection if the violation values of both individuals in $\mathbf{X}^g$ and $\mathbf{W}^g$ are the same. For such purpose, let 
\begin{equation} \label{eq:omega_phi}
    \mathbf{\Omega}_{\phi} = \mathbf{\Phi}_{\mathbf{X}^g} \overset{*}{=} \mathbf{\Phi}_{\mathbf{W}^{g}}.
\end{equation}
Then, from $\mathbf{\Omega}_{\phi}$, we substitute the corresponding elements in  
$\mathbf{\Phi}_{\mathbf{X}^g}$ and $\mathbf{\Phi}_{\mathbf{W}^{g}}$ with the objective function values, which is obtained by updating
\begin{equation}
    \begin{aligned}
        &\mathbf{\Phi}_{\mathbf{X}^g} \leftarrow \mathbf{\Phi}_{\mathbf{X}^g} \circ \overline{\mathbf{\Omega}}_{\phi} + \mathbf{f}_e(\mathbf{X}^g) \circ \mathbf{\Omega}_{\phi},\\
        &\mathbf{\Phi}_{\mathbf{W}^{g}} \leftarrow \mathbf{\Phi}_{\mathbf{W}^{g}} \circ \overline{\mathbf{\Omega}}_{\phi} + \mathbf{f}_e(\mathbf{W}^{g}) \circ \mathbf{\Omega}_{\phi},
    \end{aligned}
\end{equation}
where $\overline{\mathbf{\Omega}}_{\phi}$ is the reverse Boolean matrices of ${\mathbf{\Omega}}_{\phi}$. Consequently, the $(g+1)$th population matrix is constructed as
\begin{equation}\label{eq:update_pop_modified}
    \mathbf{X}^{g+1} = \mathbf{W}^{g} \circ \left(\mathbf{\Omega}_{E'} \mathbf{J}_{1\times D}\right) + \mathbf{X}^{g} \circ \left(\overline{\mathbf{\Omega}}_{E'} \mathbf{J}_{1\times D}\right),
\end{equation}
where
\begin{equation}
    \mathbf{\Omega}_{E'} = \mathbf{\Phi}_{\mathbf{W}^{g}} \overset{*}{<} \mathbf{\Phi}_{\mathbf{X}^g},
\end{equation}
and $\overline{\mathbf{\Omega}}_{E'}$ is the reverse Boolean matrix of $\mathbf{\Omega}_{E'}$.

\subsubsection{Algorithm summary}
% \textcolor{red}{Based on the modified selection procedure, which considers constraints as described above, the feasibility law \cite{deb2000constraint} guiding the evolution can be efficiently implemented using matrix operations. Specifically, calculations can be parallelized at the population level in actual programming.
% For convenience in comparing individuals in the population, the high-level API for the feasibility law has been provided by the Geatpy library \cite{jazzbin2020geatpy, liu2022hybridization}.
% It is important to note that $\phi(\mathbf{x})=0$ for a feasible $\mathbf{x}$.
% Therefore, if both individuals in $\mathbf{X}^g$ and $\mathbf{W}^g$ are feasible, the original objective values are used for selection, as shown in \eqref{eq:omega_phi} to \eqref{eq:update_pop_modified}. Hence, the modified selection procedure utilizes the following feasibility law:
% \begin{itemize}
%     \item Feasible individuals are accepted over infeasible individuals.
%     \item If both individuals are feasible, the one with 
% a larger objective value is chosen.
%     \item If both individuals are infeasible, the one with a smaller violation value is chosen.
% \end{itemize}
% }

Algorithm~\ref{alg:procedure} summarizes the procedure of the proposed MDE-CH for solving the optimization problem in Section~\ref{subsec:revised_opt}.
\begin{algorithm}
\caption{Proposed MDE-CH optimization}
\label{alg:procedure}
\begin{algorithmic}[1]
%\REQUIRE Problem Parameters, Matrix-based DE Parameters
\STATE{Initialize $\tau = 0$ and $\Psi = \emptyset$.}
%\STATE{{~~~}Create an empty set $\Psi = \emptyset$}
\WHILE{$\tau = 0$}
     \STATE {Population initialization: Construct $\mathbf{X}^{1}$ by performing \eqref{eq: upper_lower_vector} and \eqref{eq:ini_pop}.}
    %\STATE {Initialize population matrix: \eqref{eq:ini_pop} to obtain $\mathbf{X}^{1}$}
    \FOR {$g\in[1:G-1]$}
        \STATE {Mutation: Construct $\mathbf{Y}^{g}$ by performing \eqref{eq:amp_fac} to \eqref{eq: repair_ul}}.
        \STATE {Crossover: Construct $\mathbf{W}^{g}$ by performing \eqref{eq:cro_rate} to \eqref{eq:w_g}}.
        \STATE {Selection: Construct $\mathbf{X}^{g+1}$ by performing the selection procedure in Section \ref{subsec:seleciton_constraints}.}
    \ENDFOR
    
    \FOR{$i\in[1:N]$}
       \IF{$\phi\left(\mathbf{X}^{G}_{[i,\cdot]}\right) = 0$}
            \STATE {Update $\tau \leftarrow \tau + 1$.}
            \STATE {Update $\Psi \leftarrow \Psi \cup \{i\}$.}
       \ENDIF
    \ENDFOR
\ENDWHILE
\STATE{Initialize $f^{\operatorname{opt}} = +\infty$ and $\mathbf{x}^{\operatorname{opt}} = \mathbf{0}_{1\times D}$.}
%\STATE{Let $\mathbf{x}^{\operatorname{opt}} = \mathbf{0}_{1\times D}$}
\FOR{$j\in\Psi$}
\STATE{Evaluate the fitness value of $f_e \big(\mathbf{X}^{G}_{[j, \cdot]}\big)$ based on ${\bar{E}_{{\mathop{\rm fly}\nolimits} }} + {\bar{E}_{{\mathop{\rm com}\nolimits} }}$.}
    \IF{$f_e \big(\mathbf{X}^{G}_{[j, \cdot]}\big) < f^{\operatorname{opt}}$}
        \STATE{Update $f^{\operatorname{opt}} \leftarrow f_e \big(\mathbf{X}^{G}_{[j, \cdot]}\big)$.}
        \STATE{Update $\mathbf{x}^{\operatorname{opt}} \leftarrow \mathbf{X}^{G}_{[j, \cdot]}$.}
    \ENDIF
\ENDFOR
\STATE{Construct $\{\mathbf{b}^{\operatorname{opt}}(\bar{t})\}_{\bar{t}\in[0,1]}$ from $\mathbf{x}^{\operatorname{opt}}_{[1]}$ to $\mathbf{x}^{\operatorname{opt}}_{[D-1]}$ according to \eqref{eq:bu}.}

\STATE{Construct $\{\mathbf{q}^{\operatorname{opt}}(t)\}_{t\in[1:T^{\operatorname{opt}}]}$ from $\{\mathbf{b}^{\operatorname{opt}}(\bar{t})\}_{\bar{t}\in[0,1]}$ according to \eqref{eq:b2q}, where $T^{\operatorname{opt}} = \mathbf{x}^{\operatorname{opt}}_{[D]}$.}
%\STATE{$\mathbf{q}(t) \leftarrow \mathbf{b}\left(\frac{t}{T}\right)$, where $T = \mathbf{x}^{\operatorname{opt}}_{[1,D]}$, according to \eqref{eq:b2q}}
\ENSURE UAV Trajectory: $\{\mathbf{q}^{\operatorname{opt}}(t)\}_{t\in[1:T^{\operatorname{opt}}]}$.
\end{algorithmic}
\end{algorithm}

\section{Numerical Evaluation}\label{sec:simulation}
In this section, we numerically evaluate the performance of the proposed MDE-CH algorithm.

\subsection{Simulation Environment}
\subsubsection{3D terrain representation}
There exist several works studying 3D terrain representation. In~\cite{kamyar2014aircraft}, 3D terrain has been constructed based on randomly generated Gaussian functions. In~\cite{nikolos2003evolutionary}, functions consisting of sine and cosine have been used to describe the wave landscape. Additionally, in the field of 2D/3D map representation, cylindrical danger zones (or no-fly zones) can be formulated by inequality, such as ${(x-x_c)}^2 + {(y-y_c)}^2 \geq r^2$,
which denotes that the trajectory cannot enter the circle region, locating $(x_c, y_c)$ with a radius of $r$ \cite{roberge2012comparison}.

\begin{figure}[t!]
    \centering
    \includegraphics[width=0.5\linewidth]{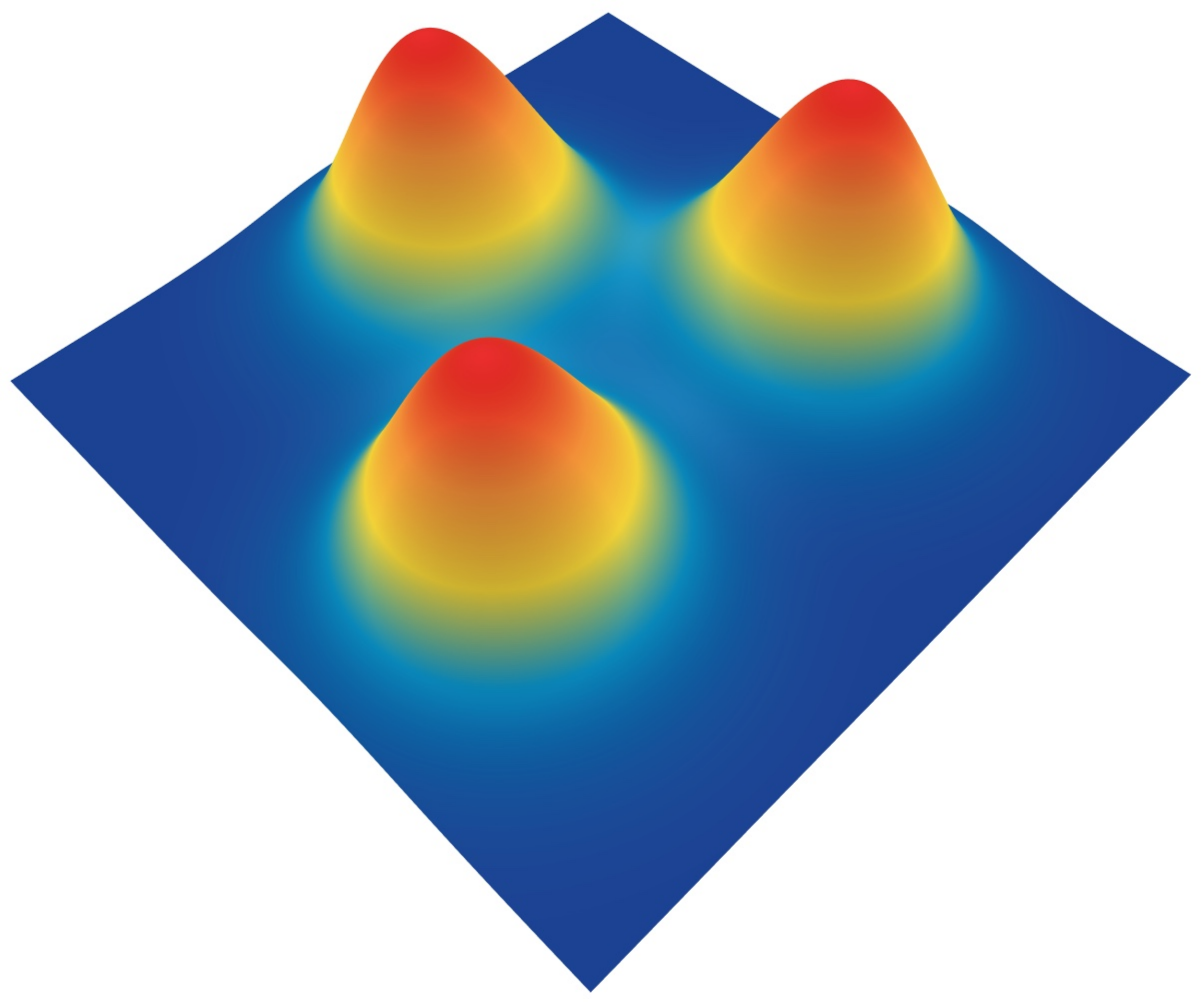}
    \caption{3D terrain generated by three Gaussian sub-functions.}
    \label{fig:map}
\end{figure}

In this paper, 3D terrain is generated by using a combination of Gaussian functions for numerical evaluation~\cite{nikolos2003evolutionary,kamyar2014aircraft,choi2018three}. Specifically, 3D terrain is represented by 
\begin{equation}\label{eq:map}
    f_z(x,y) = \sum_{i=1}^{I} A_i e^{-\left(\frac{{(x-\mu_{x_i})}^2}{2{\sigma_{x_i}^2}} + 
    \frac{{(y-\mu_{y_i})}^2}{2{\sigma_{y_i}^2}}\right)},
\end{equation}
where $f_z (x,y)$ is the ground altitude, and $\mu_{x_i}$, $\mu_{y_i}$, ${\sigma_{x_i}}$, and ${\sigma_{y_i}}$ are the parameters of each sub-function, $A_i$ is the altitude of sub-functions, and $I$ is the number of sub-functions. 
% \textcolor{blue}{Besides, the ranges of the map are $x\in [L_{x}^{*}, U_{x}^{*}]$, $y\in [L_{y}^{*}, U_{y}^{*}]$. For unified notation, we define the upper flying altitude and lower flying altitude as $[L_{z}^{*}, U_{z}^{*}]$.}
It is notable that the real-world landscape can be simulated with \eqref{eq:map} to fit the contour map with altitude using methods like the least square method or other functions, such as a neural network~\cite{elbrachter2021deep}. In the simulation, three Gaussian sub-functions, i.e., $I=3$, are employed with the parameters in Table~\ref{tab:map_parameters} and Fig.~\ref{fig:map} illustrates the resulting 3D terrain based on the parameters presented in Table~\ref{tab:map_parameters}.

\begin{table}[t!]
\caption{Parameters for 3D Terrain.}\label{tab:map_parameters}
\centering
\setlength{\tabcolsep}{4.5mm}{
\begin{tabular}{llllll}
\toprule
        & $A_i$ &$\mu_{x_i}$  & $\mu_{y_i}$ & $\sigma_{x_i}$ & $\sigma_{y_i}$ \\ \midrule
$i=1$   & 150    &200          & 500         & 90             & 90          \\
$i=2$   & 150    &600          & 500         & 90             & 90          \\
$i=3$   & 150    &400          & 200         & 90             & 90          \\
\bottomrule
\end{tabular}}
\end{table}

\subsubsection{Main simulation parameters}
%Our codes were run in a MacBook Air with an Apple M2 chip, 16 GB memory, and a macOS Sonoma 14.0 system. 

\begin{table}
\caption{Main simulation parameters.}\label{tab:parameters}
\centering
\setlength{\tabcolsep}{1mm}{
\begin{tabular}{lll}
\toprule
Notation           & Values                 & Description                   \\ \midrule
$M$              & 11                     & Number of control points          \\
$n$                & 100                    & Number of interpolation \big($n = \frac{T}{\Delta_{t}}+1$\big)        \\ \midrule
$U_{x}$          & 800 $m$                    & Maximum bound of the $x$-axis \\
$U_{y}$          & 800 $m$                  & Maximum bound of the $y$-axis \\ 
$U_{z}$          & 122 $m$                     & Maximum bound of the $z$-axis \\
$\Delta_z$         & 0.5 $m$                   & Minimum UAV distance from the ground  \\ \midrule
$W$                & 20 Newton                     & Aircraft weight     \\ 
$\rho$              & 1.225 $kg/m^3$        & Air density          \\ 
$\zeta$             & 0.4 $m$                & Rotor radius                  \\
$A$                & 0.503 $m^2$            & Rotor disc area                  \\
$\Omega$            & 300 $rad/s$           & Blade angular velocity    \\
$U_{\operatorname{tip}}$  & 120 $m/s$        & Tip speed of the rotor blade  \\
$s$                & 0.05                   & Rotor solidity                  \\
$d_0$              & 0.6                    & Fuselage drag ratio            \\
$l$                & 0.1                    & Induced power correction factor   \\
$v_0$              & 4.03 $m/s$             & Mean rotor induced velocity in hover\\
$\delta$            & 0.012                 & Profile drag coefficient    \\
$P_2$              & 11.46              & Vertical power coefficient \\ 
$v_{\operatorname{max}}$&30 $m/s$             & Maximum speed\\
$a_{i,\operatorname{max}}$& 2 $m/s^2$        & Maximum acceleration\\
\midrule
$T_{\operatorname{max}}$& 500 $s$                 & Maximum mission duration      \\
$R_{k}^{\operatorname{th}}$      & 1 Mbit/s   & Minimum data rate for communication \\
$Q_{k}^{\operatorname{th}}$    & 40-200 Mbit & Data requirement for GN $k$ \\
$P_{\operatorname{com}}$       &   5 Watt     &  Power consumption for communication \\
% $P_{t}$            & 20 dBm                  & Transmitter Power             \\
$B$                & 1 MHz                 & Bandwidth                     \\
$a$               &   10                   & LoS probability parameter 1     \\
$b$               &   0.6                  & LoS probability parameter 2     \\
$\kappa $          &   0.2                  & NLoS attenuation factor       \\
$\alpha$           &   2.3                  & Path loss exponent              \\
$\gamma_0$ &   52.5 dB              & Reference SNR             \\
\midrule
$\mathbf{p}_{1}$   & $[0,0,100]$               & Start point of the UAV trajectory ($\mathbf{q}_{\operatorname{str}}$)  \\
$\mathbf{p}_{M}$   & $[800,800,100]$           & End point of the UAV trajectory ($\mathbf{q}_{\operatorname{end}}$) \\
$K$                & 3                       & Number of GNs          \\
$\mathbf{u}_{1}$   & $[200,200,0]$            & Position of GN 1              \\
$\mathbf{u}_{2}$   & $[600,200,0]$            & Position of GN 2         \\
$\mathbf{u}_{3}$   & $[400,700,0]$            & Position of GN 3              \\\midrule
$N$                & 20                    & Population size    \\
$D$                & 28                     & Dimension size\\
$G$                & 2000                   & Maximum number of generations\\
$\lambda$                & 0.1              & Amplification factor\\
$\gamma$       & 0.5                    & Crossover rate\\
$\mathbf{U}$       & ${\{800\}}^{18}{\{122\}}^{9} \{500\}$ & Upper bounds for individuals\\
$\mathbf{L}$       & ${\{0\}}^{28}$         & Lower bounds for individuals\\
$\omega_i$         & $1/6$          & Weighting coefficients for violation\\
\bottomrule
\end{tabular}}
\end{table}

Table~\ref{tab:parameters} summarizes the main parameters for the B\'{e}zier function, 3D network area, UAV characteristics, wireless channel characteristics, sensor deployments including the start and end points of the UAV, and the proposed MDE-CH. %The detailed parameters of the map are shown in Table~\ref{tab:map_parameters}. Besides terrain and MDE-CH, all parameters can also be found in \cite{zeng2019energy}. Moreover, $P_2$ is from \cite{mei20223d}. 
Note that the flight altitude ceiling for UAVs is regulated by the Federal Aviation Administration (FAA). According to the latest regulation \cite{flying_max_altitude}, the maximum flight altitude is $400$ feet. Therefore, we set $U_z=122$ meters as an upper limit.
%In particular, we set GN 1 and 2 positions at the two sides of the third unimodal to compare the scenario in which no terrain exists. The individual dimension of MDE-CH is calculated by $(M-2) \times 3 + 1$. The upper bound $\mathbf{U}$ of individuals is related to the permutation of decision variables that have no affection for optimization performance.

\subsection{Numerical Results}
% \begin{figure*}
%     \centering
%     \includegraphics[width=0.85\linewidth]{fig3/route.pdf}
%     \caption{Optimized trajectories with obstacle detection shown in 3D viewing angle.}
%     \label{fig:3D_Route}
% \end{figure*}
% \begin{figure}
%     \centering
%     \includegraphics[width=0.8\linewidth]{fig3/route_1.pdf}
%     \caption{Optimized trajectories without obstacle detection shown in 3D viewing angle.}
%     \label{fig:3D_Route_No_Obstacle}
% \end{figure}
\begin{figure}
  \centering
  \subfigure[Trajectories with 3D terrain.]{
    \includegraphics[width=0.45\linewidth]{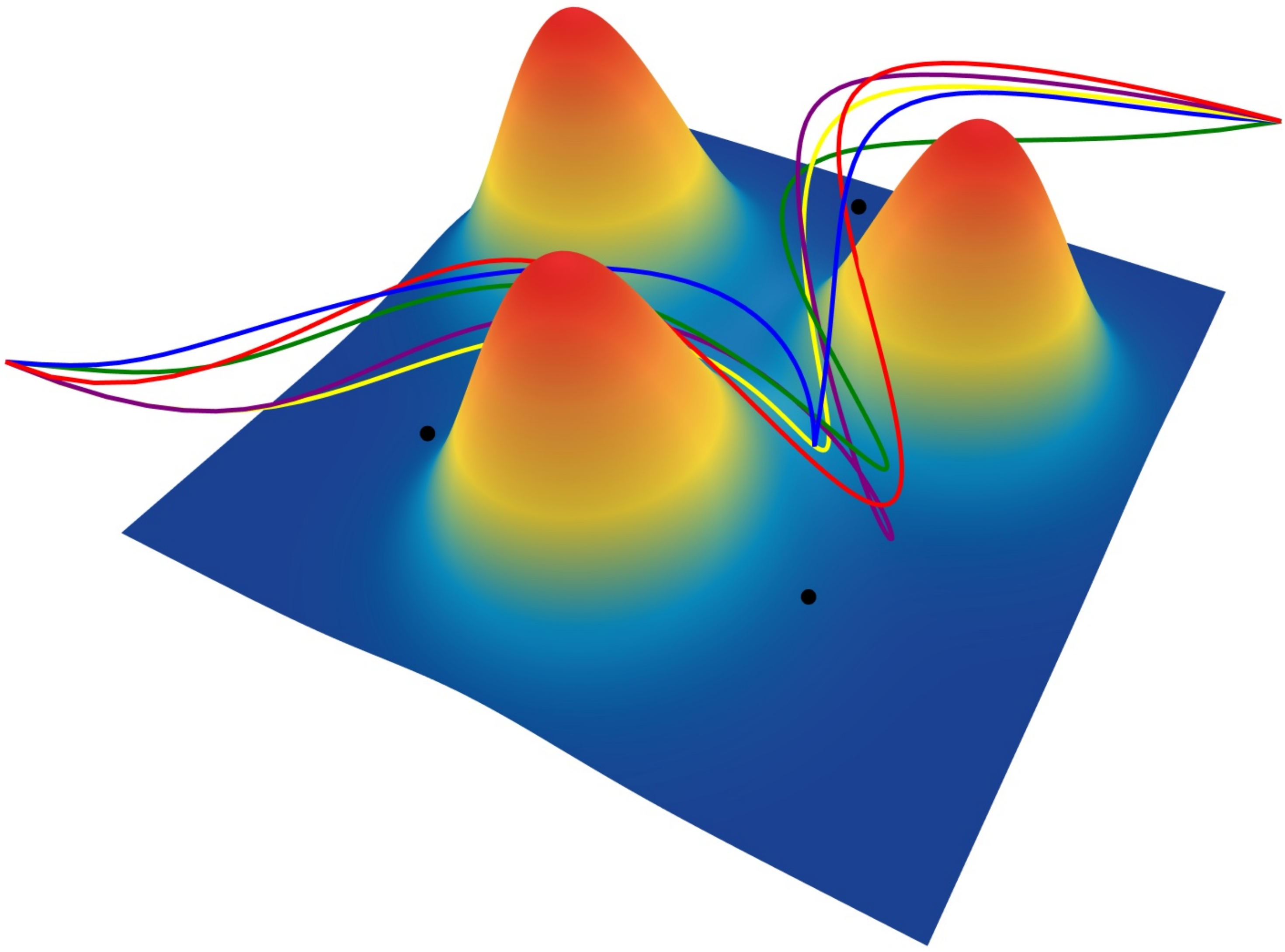}
    \label{fig:3D_Route}
  }
  \subfigure[Trajectories without 3D terrain.]{
    \includegraphics[width=0.45\linewidth]{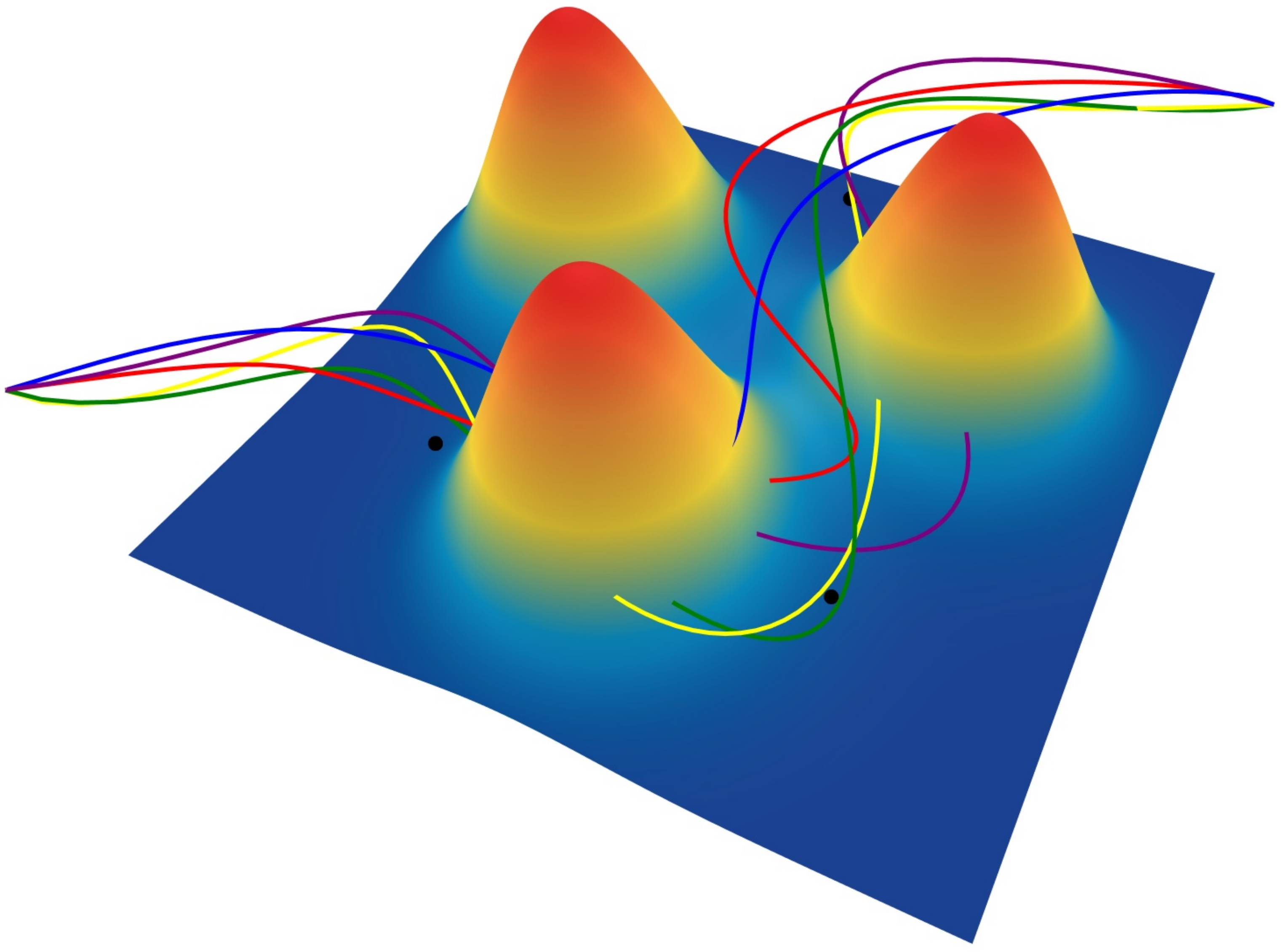}
    \label{fig:3D_Route_No_Obstacle}
  }
  \caption{Optimized UAV trajectories with and without 3D terrain.}
\end{figure}
In the simulation, we assume that the total amount of required data for all GNs is identical, i.e., $Q_{1}^{\rm{th}} = Q_{2}^{\rm{th}}=Q_{3}^{\rm{th}}$ and we uniformly use $Q_{k}^{\rm{th}}$ to denote the required data for all GNs. Similarly, we assume the identical data rate required for successful communication for all GNs, i.e., $R_{1}^{\operatorname{th}} = R_{2}^{\operatorname{th}}=R_{3}^{\operatorname{th}}$ and we uniformly use $R_{k}^{\rm{th}}$ to denote them.
Then, we optimize UAV trajectories for $Q_{k}^{\rm{th}}\in\{40, 80, 120, 160, 200\}$ Mbit.

\subsubsection{Optimization under 3D terrain}
First, we compare the proposed MDE-CH method with and without 3D terrain, i.e., the optimization problem in Section \ref{subsec:revised_opt} with and without the constraint C4 in \eqref{eq:constraint_re}.
Fig.~\ref{fig:3D_Route} plots UAV trajectories considering 3D terrain, where the black points represent the positions of GNs. It is observed that all UAV trajectories smoothly navigate around the 3D terrain. Moreover, trajectories optimized under the 3D terrain consideration exhibit more significant changes in proximity to GNs compared to those optimized without 3D terrain constraints, as depicted in Fig.~\ref{fig:3D_Route_No_Obstacle}. This is because trajectories generated with no terrain in Fig.~\ref{fig:3D_Route_No_Obstacle} are prone to go through 3D terrain to reduce energy consumption compared with trajectories considering 3D terrain. Thus, curvatures in UAV trajectories with 3D terrain are larger than those without 3D terrain, which corresponds to the results obtained by solving more complicated optimization problems. 
\begin{figure*}
  \centering
  \subfigure[UAV trajectories with different $Q_{k}^{\operatorname{th}}$.]{
    \includegraphics[width=0.28\textwidth]{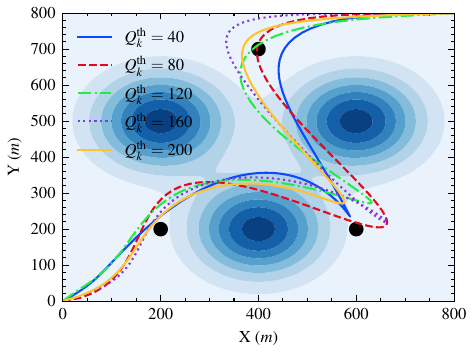}
    \label{fig:subfig1_2D_trja}
  }
  \subfigure[UAV flying speed versus time.]{
    \includegraphics[width=0.28\textwidth]{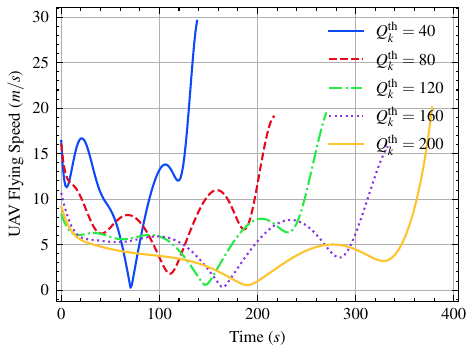}
    \label{fig:subfig2_UAV_speed}
  }
    \subfigure[UAV flying altitude versus time.]{
    \includegraphics[width=0.28\textwidth]{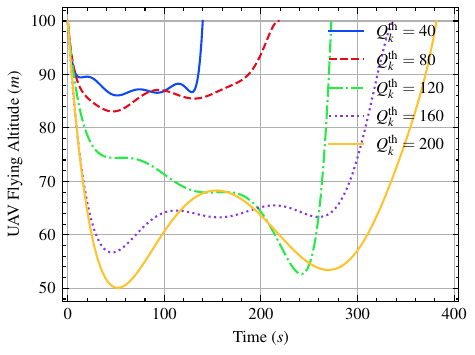}
    \label{fig:subfig3_UAV_altitude}
  }
  \caption{UAV Trajectories, flying speed, and flying altitude for different $Q_{k}^{\operatorname{th}}$.}
  \label{fig:contour_speed_altitude}
\end{figure*}
\begin{figure}
  \centering
  \subfigure[$Q_{k}^{\rm{th}} = 40$ Mbits.]{
    \includegraphics[width=0.45\linewidth]{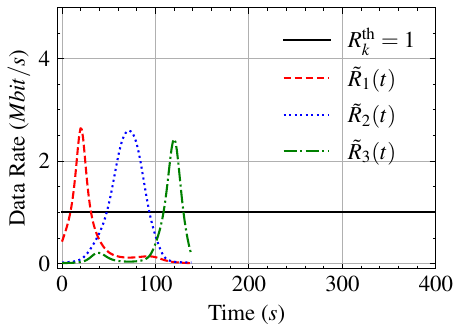}
    \label{fig:subfig1_traj_dr}
  }
    \subfigure[$Q_{k}^{\rm{th}} = 200$ Mbits.]{
    \includegraphics[width=0.45\linewidth]{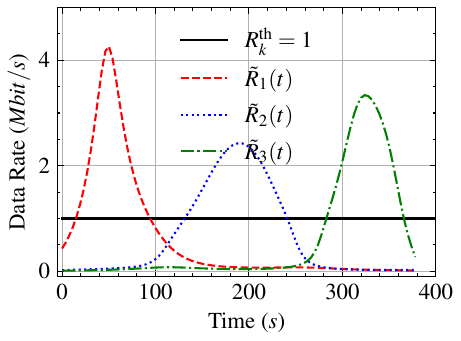}
    \label{fig:subfig5_traj_dr}
  }
  \caption{Instantaneous data rates over time between the UAV and each GN.}
  \label{fig:traj_dr}
\end{figure}

\subsubsection{UAV trajectories for different data requirement}
Fig. \ref{fig:subfig1_2D_trja} illustrates the 2D viewing angle derived from the 3D view shown in Fig. \ref{fig:3D_Route}. Also, Fig.~\ref{fig:subfig2_UAV_speed} and  Fig.~\ref{fig:subfig3_UAV_altitude} plot the UAV flying speed and the UAV altitude versus time, respectively. All curves in Fig.~\ref{fig:contour_speed_altitude} exhibit spatial and temporal smoothness, demonstrating that the proposed trajectory optimization efficiently ensures continuous UAV trajectories optimized over a continuous domain.
% The shown trajectories will be close to GN to complete the mission of communication. The greater the amount of communication throughput required, the closer the trajectory is to the communication point. 
Some trajectories in Fig.~\ref{fig:subfig1_2D_trja} exhibit high curvature regimes near GN 2, particularly those with $Q_{k}^{\operatorname{th}}=40$ Mbit and $Q_{k}^{\operatorname{th}}=160$ Mbit. These correspond to the regimes where the UAV flying speed decreases to approximately zero as observed in Fig.~\ref{fig:subfig2_UAV_speed}. This phenomenon demonstrates the advantage of using the B\'{e}zier curve in representing trajectories, aligning well with the physical constraints of UAV flight. Furthermore, the UAV reduces its speed as it approaches each GN to efficiently collect data. Additionally, as shown in Fig.~\ref{fig:subfig2_UAV_speed} and Fig.~\ref{fig:subfig3_UAV_altitude}, the mission completion time increases with higher data requirements $Q_{k}^{\operatorname{th}}$, since both the average UAV speed and altitude are reduced to meet these requirements.

%Moreover, another interesting phenomenon is the UAV will elevate the speed to reduce time consumption and further reduce energy consumption when flying to the endpoint after finishing all communication missions, even reaching the maximum speed when $Q_{k}^{\operatorname{th}} = 40$. For the results of Fig.~\ref{fig:subfig3_UAV_altitude}, the flying altitude optimization is important, which is different from 2D trajectory design, since the trajectory can find optimal communication altitude which influences the elevation angle in \eqref{eq:elevation_angle}. We can see from Fig.~\ref{fig:subfig3_UAV_altitude} that the altitude of UAV fluctuates due to finding the optimal communication position in 3D space.

Fig.~\ref{fig:traj_dr} plots instantaneous data rates between the UAV and each GN when  $Q_{k}^{\operatorname{th}}=40$ Mbit and $Q_{k}^{\operatorname{th}}=200$ Mbit, corresponding to the trajectories shown in Fig.~\ref{fig:subfig1_2D_trja}. The black straight line in each subfigure represents the minimum data rate for successful communication, namely $R_{k}^{\operatorname{th}}=1$ Mbit/s. 
The two intersections between $R_{k}^{\operatorname{th}}$ and $\tilde{R}_k(t)$ indicate the start and end timings of communication between the UAV and GN $k$, respectively. Compared with the conventional fly--hover--fly model~\cite{zeng2019energy, zhong2018secure, yao2020joint}, our model allows the UAV to gather data while it is moving, providing a more flexible design of UAV trajectories and a more efficient method for data transmission and reception.
%The other situation of remaining $\mathrm{Q}^{th}_{k}$ can be checked in \cite{extra_results}.
% \begin{figure}
%     \centering
%     \includegraphics[width=0.6\linewidth]{fig3/penalty_comparison.pdf}
%     \caption{The comparison of MDE-CH among MDE-Penalty, GA-Penalty, ES-Penalty.}
%     \label{fig:comparison_CH_Penalty}
% \end{figure}
\subsubsection{Convergence behavior}
\begin{figure}
  \centering
  \subfigure[Average value of objective functions.]{
    \includegraphics[width=0.45\linewidth]{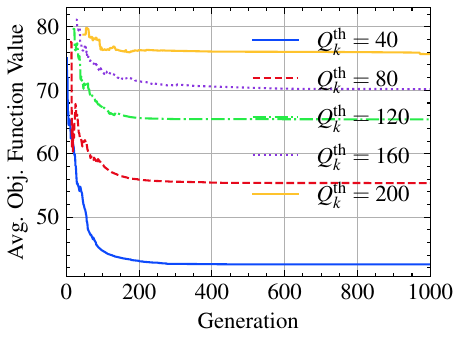}
    \label{fig:obj_fun_opt}
  }
    \subfigure[Average number of feasible solutions.]{
    \includegraphics[width=0.45\linewidth]{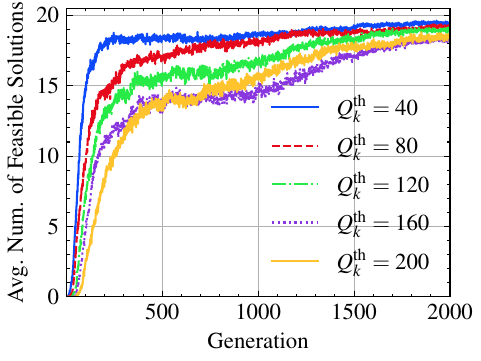}
    \label{fig:MDE-CH_FN}
    }
  \caption{Convergence behavior of the proposed MDE-CH with respect to the generation number.}
  \label{fig:opt_process}
\end{figure}
Fig.~\ref{fig:opt_process} shows the convergence behavior of the proposed MDE-CH with respect to the generation number for $Q_{k}^{\rm{th}}\in\{40, 80, 120, 160, 200\}$ Mbit.
In the simulations, each scenario was executed $50$ times  to obtain average values of the objective functions and the count of feasible solutions. 
As seen in Fig.~\ref{fig:obj_fun_opt}, the objective function rapidly converges to a stable value as the generation number increases. Fig.~\ref{fig:MDE-CH_FN} plots the average number of feasible solutions with respect to the generation number. Due to the random generation of initial individuals, only a few are feasible in the early generations. However, as the evolutionary process progresses, the population quickly transitions into the feasible region, with an increasing appearance of feasible individuals. Therefore, the proposed methodology for handling constraints effectively finds feasible solutions for the constrained optimization problem.

\subsubsection{Comparison with the fly--hover--fly model}
\begin{figure}
    \centering
    \includegraphics[width=0.6\linewidth]{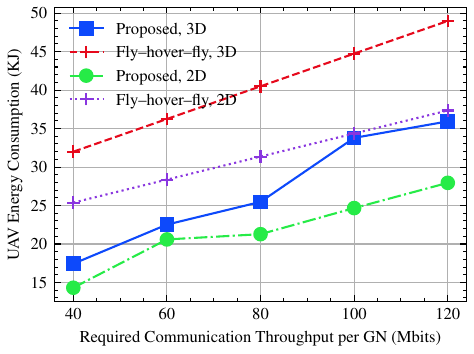}
    \caption{Energy consumption for the proposed MDE-CH and fly--hover--fly schemes.}
    \label{fig:comparison}
\end{figure}

We compare our scheme with the fly--hover--fly scheme, as described in \cite{zeng2019energy, zhong2018secure, yao2020joint}. In the fly--hover--fly model, the UAV travels directly from one GN to another at maximum speed, subsequently hovering over the GN until data reception is completed. After the reception, the UAV swiftly moves to the next GN at maximum speed. In the fly--hover--fly scenario, the UAV moves vertically for data reception at each GN. In contrast, our model requires continuous vertical and horizontal navigation of the UAV to circumvent 3D terrain obstacles.

Fig.~\ref{fig:comparison} illustrates the overall energy consumption of the proposed and fly--hover--fly schemes for $Q_{k}^{\rm{th}}\in\{40, 60, 80, 100, 120\}$ Mbit.
In the figure, ``Proposed, 3D" and ``Fly--hover--fly, 3D" denote the proposed scheme and fly--hover--fly scheme considering 3D terrain, respectively. To further compare the performance between two schemes, we also consider the 2D scenario, i.e., $f_z(u_{k,x}, u_{k,y})=0$ in \eqref{eq:u_k}, corresponding to ``Proposed, 2D" and ``Fly--hover--fly, 2D". The numerical results demonstrate that the proposed MDE-CH scheme outperforms the fly--hover--fly scheme in both 3D and 2D UAV trajectory optimization scenarios.   

\subsubsection{Comparison with other algorithms}
\begin{figure}
    \centering
    \includegraphics[width=0.6\linewidth]{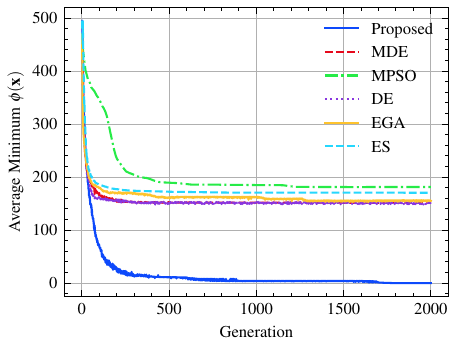}
    \caption{The results of the comparison with some EC algorithms.}
    \label{fig:comparison_penalty}
\end{figure}
Moreover, we compare the proposed MDE-CH with other evolutionary computation algorithms, including MDE, matrix-based particle swarm optimization (MPSO) \cite{zhan2021matrix}, DE \cite{ahmad2022differential}, elitism genetic algorithm (EGA) \cite{katoch2021review}, and evolution strategy (ES) \cite{beyer2002evolution}. We implemented the benchmark algorithms by applying the penalty function method to validate the efficiency of the proposed constraints handle method. The penalty function method transforms the constraints violation value $\phi(\mathbf{x})$ into the objective function as 
\begin{equation}
     \min\limits_{\mathbf{x}} f(\mathbf{x}) + \tau \phi^{2}(\mathbf{x}),\\
 \end{equation}
where $\tau$ is a coefficient to balance the constraint violation value and objective function value. In the simulation, we set $\tau=1$.
Fig. \ref{fig:comparison_penalty} illustrates the comparison results obtained by recording the minimum constraint violation value in each generation under $Q^{\mathrm{th}}_{k}=200$. We repeated the experiment $50$ times to obtain the average value. The results demonstrate that the proposed MDE-CH can find feasible solutions as the number of generations increases,  while other benchmark algorithms tend to produce solutions that fall outside the feasible region.

% In addition to the aforementioned analysis, the model effectively addresses a traveling salesman problem (TSP), specifically regarding the communication order to each GN. Leveraging the exceptional performance of MDE-CH, the communication order can be implicitly optimized by minimizing the total energy consumption of the UAV in our model.

\section{Conclusion}\label{sec:conclusion}
In this paper, we considered UAV-assisted IoT data collection in which a UAV traverses to collect sensing data from several GNs aimed at minimizing total energy consumption while accounting for the UAV's physical capabilities, the heterogeneous data demands of GNs, and 3D terrain.
We proposed the MDE-CH algorithm, a computation-efficient evolutionary algorithm designed to address non-convex constrained optimization problems with several different types of constraints that can provide a continuous 3D temporal--spatial UAV trajectory capable of efficiently minimizing energy consumption under various practical constraints.

As a future research topic, an integrated framework for optimizing both UAV and low-earth-orbit or very-low-earth-orbit satellite trajectories in sixth-generation non-terrestrial networks could be studied. In particular, the heterogeneous characteristics and requirements of various non-terrestrial communication objects should be appropriately considered.

% \appendices

% \section{}

% % use section* for acknowledgment
% \section*{Acknowledgment}


\bibliographystyle{IEEEtran}
\bibliography{IEEEabrv,ref2}

\ifCLASSOPTIONcaptionsoff
  \newpage
\fi
\end{document}